\pdfoutput=1

\documentclass[twocolumn,journal]{IEEEtran}

\usepackage[T1]{fontenc}
\usepackage[utf8]{inputenc}
\usepackage{color}
\usepackage{array}
\usepackage{booktabs}
\usepackage{units}
\usepackage{url}
\usepackage{amsmath}
\usepackage{amssymb}
\usepackage{graphicx}
\usepackage{textcomp}
\PassOptionsToPackage{normalem}{ulem}
\usepackage{ulem}
\usepackage[unicode=true,
 bookmarks=true,bookmarksnumbered=true,bookmarksopen=true,bookmarksopenlevel=1,
 breaklinks=false,pdfborder={0 0 0},pdfborderstyle={},backref=false,colorlinks=false]
 {hyperref}
\hypersetup{pdftitle={Your Title},
 pdfauthor={Your Name},
 pdfpagelayout=OneColumn, pdfnewwindow=true, pdfstartview=XYZ, plainpages=false}

\makeatletter

\providecommand{\tabularnewline}{\\}

\let\oldforeign@language\foreign@language
\DeclareRobustCommand{\foreign@language}[1]{%
  \lowercase{\oldforeign@language{#1}}}

\usepackage[caption=false,font=footnotesize]{subfig}
\usepackage{xcolor}
\usepackage{graphicx}
\usepackage{cite}
\usepackage{balance}

\@ifundefined{showcaptionsetup}{}{%
 \PassOptionsToPackage{caption=false}{subfig}}
\usepackage{subfig}
\makeatother

\usepackage{listings}
\begin{document}
\title{DQ Robotics: a Library for Robot Modeling and Control}
\author{Bruno~Vilhena~Adorno,~\IEEEmembership{Senior Member,~IEEE} and
Murilo~Marques~Marinho,~\IEEEmembership{Member,~IEEE}\thanks{This work was supported by CNPq, CAPES, FAPEMIG.}\thanks{B.V. Adorno is with the Federal University of Minas Gerais (UFMG),
Department of Electrical Engineering, 31270-010, Belo Horizonte-MG,
Brazil. e-mail: \protect\href{http://adorno\%40ufmg.br}{adorno@ufmg.br}.
B. V. Adorno is supported by CNPq Grant Numbers 424011/2016-6 and
303901/2018-7.}\thanks{M. M. Marinho is with the University of Tokyo (UTokyo), Department
of Mechanical Engineering, Tokyo, Japan. e-mail: \protect\href{http://murilo\%40nml.t.u-tokyo.ac.jp}{murilo@nml.t.u-tokyo.ac.jp}.
M. M. Marinho is supported by JSPS KAKENHI Grant Number 19K14935.}\thanks{\textcopyright 2020 IEEE.  Personal use of this material is permitted.  Permission from IEEE must be obtained for all other uses, in any current or future media, including reprinting/republishing this material for advertising or promotional purposes, creating new collective works, for resale or redistribution to servers or lists, or reuse of any copyrighted component of this work in other works.}}
\markboth{Accepted for publication in IEEE Robotics \& Automation Magazine}{B.V. Adorno and M. M. Marinho: DQ Robotics: a Library for Robot Modeling
and Control}
\maketitle
\begin{abstract}
Dual quaternion algebra and its application to robotics have gained
considerable interest in the last two decades. Dual quaternions have
great geometric appeal and easily capture physical phenomena inside
an algebraic framework that is useful for both robot modeling and
control. Mathematical objects, such as points, lines, planes, infinite
cylinders, spheres, coordinate systems, twists, and wrenches are all
well defined as dual quaternions. Therefore, simple operators are
used to represent those objects in different frames and operations
such as inner products and cross products are used to extract useful
geometric relationships between them. Nonetheless, the dual quaternion
algebra is not widespread as it could be, mostly because efficient
and easy-to-use computational tools are not abundant and usually are
restricted to the particular algebra of quaternions. To bridge this
gap between theory and implementation, this paper introduces DQ Robotics,
a library for robot modeling and control using dual quaternion algebra
that is easy to use and intuitive enough to be used for self-study
and education while being computationally efficient for deployment
on real applications.

\end{abstract}

\global\long\def\dq#1{\underline{\boldsymbol{#1}}}%

\global\long\def\quat#1{\boldsymbol{#1}}%

\global\long\def\mymatrix#1{\boldsymbol{#1}}%

\global\long\def\myvec#1{\boldsymbol{#1}}%

\global\long\def\mapvec#1{\boldsymbol{#1}}%

\global\long\def\dualvector#1{\underline{\boldsymbol{#1}}}%

\global\long\def\dual{\varepsilon}%

\global\long\def\dotproduct#1{\langle#1\rangle}%

\global\long\def\norm#1{\left\Vert #1\right\Vert }%

\global\long\def\mydual#1{\underline{#1}}%

\global\long\def\hamilton#1#2{\overset{#1}{\operatorname{\mymatrix H}}\left(#2\right)}%

\global\long\def\hamiquat#1#2{\overset{#1}{\operatorname{\mymatrix H}}_{4}\left(#2\right)}%

\global\long\def\hami#1{\overset{#1}{\operatorname{\mymatrix H}}}%

\global\long\def\tplus{\dq{{\cal T}}}%

\global\long\def\getp#1{\operatorname{\mathcal{P}}\left(#1\right)}%

\global\long\def\getd#1{\operatorname{\mathcal{D}}\left(#1\right)}%

\global\long\def\swap#1{\text{swap}\{#1\}}%

\global\long\def\imi{\hat{\imath}}%

\global\long\def\imj{\hat{\jmath}}%

\global\long\def\imk{\hat{k}}%

\global\long\def\real#1{\operatorname{\mathrm{Re}}\left(#1\right)}%

\global\long\def\imag#1{\operatorname{\mathrm{Im}}\left(#1\right)}%

\global\long\def\imvec{\boldsymbol{\imath}}%

\global\long\def\vector{\operatorname{vec}}%

\global\long\def\mathpzc#1{\fontmathpzc{#1}}%

\global\long\def\cost#1#2{\underset{\text{#2}}{\operatorname{\text{cost}}}\left(\ensuremath{#1}\right)}%

\global\long\def\diag#1{\operatorname{diag}\left(#1\right)}%

\global\long\def\frame#1{\mathcal{F}_{#1}}%

\global\long\def\ad#1#2{\text{Ad}\left(#1\right)#2}%

\global\long\def\adsharp#1#2{\text{Ad}_{\sharp}\left(#1\right)#2}%

\global\long\def\spin{\text{Spin}(3)}%

\global\long\def\spinr{\text{Spin}(3){\ltimes}\mathbb{R}^{3}}%

\global\long\def\argminimtwo#1#2#3#4#5{ \begin{aligned}#1\:  &  \underset{#2}{\arg\!\min}  &   &  #3 \\
  &  \text{subject to}  &   &  #4\\
  &   &   &  #5 
\end{aligned}
 }%

\section{Introduction}

\IEEEPARstart{D}{ual quaternion algebra} and its application to robotics
have gained considerable interest in the last two decades. Far from
being an abstract mathematical tool, dual quaternions have great geometric
appeal and easily capture physical phenomena inside an algebraic framework
that is useful for both robot modeling and control. Mathematical objects,
such as points, lines, planes, infinite cylinders, spheres, coordinate
systems, twists, and wrenches are all well defined as dual quaternions.
Therefore, simple operators are used to represent those objects in
different coordinate systems and operations such as inner products
and cross products are used to extract useful geometric relationship
between them. Some authors consider the particular set of dual quaternions
with unit norm, known as unit dual quaternions, as the most efficient
and compact tool to describe rigid transformations \cite{BottemaRoth:1979,McCarthy1990}.
For instance, homogeneous transformation matrices (HTM) have sixteen
elements, whereas dual quaternions have eight elements and dual quaternion
multiplications are less expensive than HTM multiplications \cite[p. 42]{Adorno2011e}.
Moreover, it is easy to extract geometric parameters from a given
unit dual quaternion (translation, axis of rotation, angle of rotation).
Nonetheless, they are easily mapped into a vector structure, which
can be particularly convenient to perform tasks such as pose control
as there is no need to extract parameters from the dual quaternion.

Nonetheless, the dual quaternion algebra is not widespread as it could
be, not only because classic matrix algebra on real numbers is very
mature and it is the backbone of most robotics textbooks \cite{spong2008robot,siciliano2010robotics,siciliano2016springer},
but also because efficient and easy-to-use computational tools are
not abundant and usually are restricted to the particular algebra
of quaternions. Indeed, the Boost Math library\footnote{https://www.boost.org/}
implements quaternions and even octonions, but not dual quaternions,
whereas the Eigen\footnote{http://eigen.tuxfamily.org/} library implements
only quaternions, both in C++ language. Some libraries also implement
dual quaternions in Lua,\footnote{https://esslab.jp/\textasciitilde ess/en/code/libdq/}
MATLAB \cite{Leclercq2013}, and C++,\footnote{https://glm.g-truc.net}
but none of them are focused on robotics.

This paper introduces DQ Robotics, a library for robot modeling and
control using dual quaternion algebra that is computationally efficient,
easy to use, and is intuitive enough to be used for self-study and
education and sufficiently efficient for deployment on real applications.
For instance, DQ Robotics has already been used in real platforms
such as cooperative manipulators for surgical applications, mobile
manipulators, and humanoids, among several other robotic systems,
some of which are shown in Fig.~\ref{fig:applications}. It is written
in three languages, namely Python, MATLAB, and C++, all of them sharing
a unified programming style to make the transition from one language
to another as smooth as possible, enabling fast prototype-to-release
cycles. Furthermore, a great effort has been made to make coding as
close as possible to the mathematical notation used on paper, making
it easy to implement code as soon as one has grasped the mathematical
concepts.

DQ Robotics uses the expressiveness of dual quaternion algebra for
both robot modeling and control. This paper introduces the main features
of the library as well as its basic usage in a tutorial-like style.
The tutorial begins with the presentation of dual quaternion notation
and basic operations, goes through robot modeling and control, and
ends with a complete robot control example of two robots cooperating
in a task where one manipulator robot and a mobile manipulator interact
while deviating from obstacles in the workspace.

\subsection{How to follow the tutorial}

In this paper, for brevity, we present several code snippets in MATLAB
language that should be familiar to most roboticists. Suitable commands
are also available in Python and C++. The readers might be interested
in downloading and installing the DQ Robotics toolbox\footnote{\url{https://github.com/dqrobotics/matlab/releases/latest}}
for MATLAB and trying out the code snippets in their own machine.
More details are given in Section~\ref{sec:development_infrastructure}
and in the DQ Robotics documentation.\footnote{\url{https://dqroboticsgithubio.readthedocs.io/en/latest/installation.html}}

\begin{figure}[tbh]
\begin{centering}
\includegraphics[width=1\columnwidth]{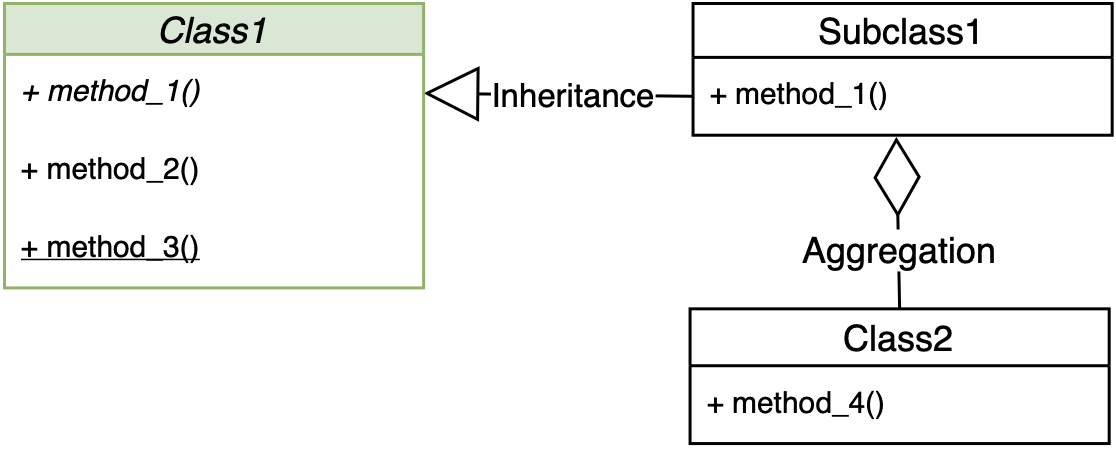}
\par\end{centering}
\caption{\label{fig:uml_explanation}A simplified class diagram in the Unified
Modeling Language (UML). Abstract methods and classes are \emph{italicized}.
Concrete methods and classes are in upright text. Static method are
\uline{underlined}. Specific arrowheads indicate inheritance and
aggregation.}
\end{figure}

The tutorial also has a few diagrams using the Unified Modeling Language
(UML), which we use to explain the object-oriented modeling of the
library in a way that is agnostic to the programming language. A simplified
UML diagram that shows the relevant object-oriented concepts used
in this paper is briefly explained in Fig.~\ref{fig:uml_explanation}.

\subsection{Notation choice and common operations}

Given the imaginary units $\imi$,$\imj$,$\imk$ that satisfy $\imi^{2}=\imj^{2}=\imk^{2}=\imi\imj\imk=-1$
and the dual unit $\dual$, which satisfies $\dual\neq0$ and $\dual^{2}=0$,
the quaternion set is defined as
\[
\mathbb{H}\triangleq\left\{ h_{1}+\imi h_{2}+\imj h_{3}+\imk h_{4}\,:\,h_{1},h_{2},h_{3},h_{4}\in\mathbb{R}\right\} 
\]
and the dual quaternion set is defined as
\[
\mathcal{H}\triangleq\left\{ \quat h_{1}+\dual\quat h_{2}\,:\,\quat h_{1},\quat h_{2}\in\mathbb{H}\right\} .
\]

There are several equivalent notations for representing dual quaternions,
but we adopt the hypercomplex one, which regards them as an extension
of complex numbers. This way, sums, multiplications and subtractions
are exactly the same used in complex numbers (and also in real numbers!),
differently from the scalar-plus-vector notation, which requires the
redefinition of the multiplication operation. When designing DQ Robotics,
one important guiding principle was to be able to use dual quaternion
operations as close as possible to the way that we do it on paper,
closing the gap between theory and implementation.

For instance, the dual quaternions $\dq a=\imi+\dual\left(1+\imk\right)$
and $\dq b=-2+\imj+\dual\left(\imi+\imk\right)$ are declared in MATLAB
as\lstset{language=Matlab}

\begin{lstlisting}
a = DQ.i + DQ.E*(1 + DQ.k);
b = -2 + DQ.j + DQ.E*(DQ.i + DQ.k);
\end{lstlisting}
and 
\begin{lstlisting}
>> a + b
ans = ( - 2 + 1i + 1j) + E*(1 + 1i + 2k)
>> a * b
ans = ( - 2i + 1k) + E*( - 3 - 1i - 2k)
\end{lstlisting}
Alternatively, one can use \texttt{include\_namespace\_dq} to enable
the aliases \texttt{i\_},\texttt{ j\_}, \texttt{k\_}, and \texttt{E\_}.
Therefore,
\begin{lstlisting}
>> E_*a
ans = E*(1i)
\end{lstlisting}

The list of the main DQ class operations is shown in Table~\ref{tab:Main-methods-of-DQ}.

\noindent 
\begin{table}[tbh]
\caption{Main methods of class \texttt{DQ}.\label{tab:Main-methods-of-DQ}}

\renewcommand*\arraystretch{1}
\noindent \begin{centering}
\begin{tabular*}{1\columnwidth}{@{\extracolsep{\fill}}>{\raggedright}p{0.25\columnwidth}>{\raggedright}p{0.7\columnwidth}}
\hline 
\multicolumn{2}{c}{\textbf{Binary operations between two dual quaternions (e.g., }\texttt{\textbf{a
+ b}}\textbf{)}}\tabularnewline
\texttt{+},\texttt{ -,{*}} & Sum, subtraction, and multiplication.\tabularnewline
\texttt{/},\texttt{ \textbackslash} & Right division and left division.\tabularnewline
\hline 
\multicolumn{2}{c}{\textbf{Unary operations (e.g., }\texttt{\textbf{-a}}\textbf{ and
}\texttt{\textbf{a'}}\textbf{)}}\tabularnewline
\texttt{-} & Minus.\tabularnewline
\texttt{{[}'{]}}, \texttt{{[}.'{]}} & Conjugate and sharp conjugate.\tabularnewline
\hline 
\multicolumn{2}{c}{\textbf{Unary operators (e.g., }\texttt{\textbf{P(a)}}\textbf{)}}\tabularnewline
\texttt{conj, sharp} & Conjugate and sharp conjugate.\tabularnewline
\texttt{exp}, \texttt{log} & Exponential of pure dual quaternions and logarithm of unit dual quaternions.\tabularnewline
\texttt{inv} & Inverse under multiplication.\tabularnewline
\texttt{hamiplus4}, \texttt{haminus4} & Hamilton operators of a quaternion.\tabularnewline
\texttt{hamiplus8}, \texttt{haminus8} & Hamilton operators of a dual quaternion.\tabularnewline
\texttt{norm} & Norm of dual quaternions.\tabularnewline
\texttt{P, D} & Primary part and dual part.\tabularnewline
\texttt{Re, Im} & Real component and imaginary components.\tabularnewline
\texttt{rotation} & Rotation component of a unit dual quaternion.\tabularnewline
\texttt{rotation\_axis}, \texttt{rotation\_angle} & Rotation axis and rotation angle of a unit dual quaternion.\tabularnewline
\texttt{vec3}, \texttt{vec6} & Three-dimensional and six-dimensional vectors composed of the coefficients
of the imaginary part of a quaternion and a dual quaternion, respectively.\tabularnewline
\texttt{vec4}, \texttt{vec8} & Four-dimensional and eight-dimensional vectors composed of the coefficients
of a quaternion and a dual quaternion, respectively.\tabularnewline
\hline 
\multicolumn{2}{c}{\textbf{Binary operators (e.g., }\texttt{\textbf{Ad(a,b)}}\textbf{)}}\tabularnewline
\texttt{Ad, Adsharp} & Adjoint operation and sharp adjoint.\tabularnewline
\texttt{cross, dot} & Cross product and dot product.\tabularnewline
\hline 
\multicolumn{2}{c}{\textbf{Binary relations (e.g. }\texttt{\textbf{a == b}}\textbf{)}}\tabularnewline
\textsf{==, \textasciitilde =} & Equal and not equal.\tabularnewline
\hline 
\multicolumn{2}{c}{\textbf{Common methods}}\tabularnewline
\texttt{{[}plot{]}} & Plots coordinate systems, lines, and planes.\tabularnewline
\hline 
\end{tabular*}
\par\end{centering}
\medskip{}

{*}Methods and operations enclosed by brackets (e.g., \texttt{{[}.{*}{]}})
are available only on MATLAB.
\end{table}

\section{Representing rigid motions}

Since the set of dual quaternions represents an eight-dimensional
space, it is particularly useful to represent rotations, translations,
and, more generally, rigid motions. For instance, the set $\mathbb{S}^{3}\triangleq\left\{ \quat h\in\mathbb{H}:\norm{\quat h}=1\right\} $
is used to represent rotations, with $\norm{\quat h}\triangleq\sqrt{\quat h\quat h^{*}}=\sqrt{h_{1}^{2}+h_{2}^{2}+h_{3}^{2}+h_{4}^{2}}$
being the quaternion norm and $\quat h^{*}=h_{1}-(h_{2}\imi+h_{3}\imj+h_{4}\imk)$
being the conjugate of $\quat h=h_{1}+h_{2}\imi+h_{3}\imj+h_{4}\imk$.
Any element $\quat r\in\mathbb{S}^{3}$ can always be written as $\quat r=\cos\left(\phi/2\right)+\quat n\sin\left(\phi/2\right)$,
where $\quat n=n_{x}\imi+n_{y}\imj+n_{z}\imk\in\mathbb{S}^{3}$ is
the unit-norm rotation axis and $\phi\in[0,2\pi)$ is the rotation
angle. Therefore, a rotation of $\pi$ rad around the $x$-axis is
given by $\quat r_{\pi,x}\triangleq\cos(\pi/2)+\imi\sin(\pi/2)=\imi.$

\begin{figure}[tbh]
\noindent \begin{centering}
\includegraphics[width=1\columnwidth]{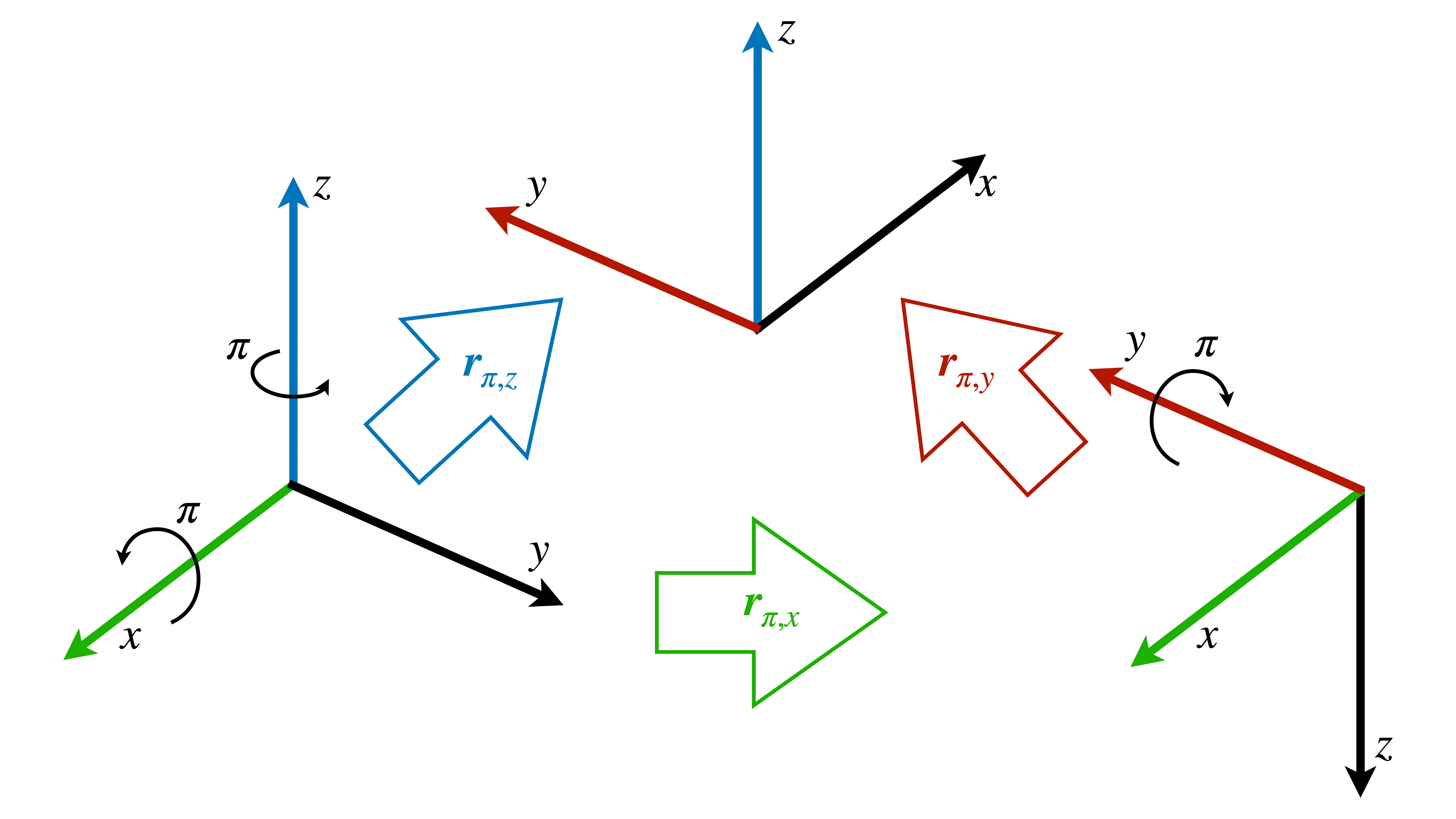}
\par\end{centering}
\caption{Sequence of rotations represented by multiplication of unit quaternions.
Each arrow has the same color of the rotation axis. \label{fig:Sequence-of-rotations}}
\end{figure}

Since the multiplication of unit quaternions is also a unit quaternion,
a sequence of rotations is given by a sequence of multiplications
of unit quaternions. For instance, if first we perform the rotation
$\quat r_{\pi,x}$ and then we use the rotated frame to do another
rotation of $\pi$ rad around the $y$-axis, given by $\quat r_{\pi,y}\triangleq\cos(\pi/2)+\imj\sin(\pi/2)=\imj$,
then the final rotation is given by $\quat r_{\pi,x}\quat r_{\pi,y}=\imi\imj=\imk$
because $\imi\imj\imk=-1$ and $\imk^{2}=-1$. Besides, because $\imk=\cos(\pi/2)+\imk\sin(\pi/2)$,
then $\quat r_{\pi,x}\quat r_{\pi,y}=\quat r_{\pi,z}$, which represents
the rotation of $\pi$~rad around the $z$-axis. This is illustrated
in Fig.~\ref{fig:Sequence-of-rotations}.

The set of pure quaternions is defined as $\mathbb{H}_{p}\triangleq\left\{ h_{1}\imi+h_{2}\imj+h_{3}\imk:h_{1},h_{2},h_{3}\in\mathbb{R}\right\} \subset\mathbb{H}$
and is used to represent elements of a tridimensional space. For instance,
the point $\left(x,y,z\right)$ can be directly mapped to $x\imi+y\imj+z\imk$,
which is an element of $\mathbb{H}_{p}$. Furthermore, given a point
$\quat p^{0}\in\mathbb{H}_{p}$ in frame $\frame 0$ and the unit
quaternion $\quat r_{1}\in\mathbb{S}^{3}$ that represents the rotation
from $\frame 0$ to $\frame 1$, the coordinates of the point in $\mathcal{F}_{1}$
is $\quat p^{1}=\quat r_{1}^{*}\quat p^{0}\quat r_{1}$. Analogously,
given $\quat r_{2}\in\mathbb{S}^{3}$ that represents the rotation
from $\frame 1$ to $\frame 2$, the coordinates of the point in $\frame 2$
is $\quat p^{2}=\quat r_{2}^{*}\quat p^{1}\quat r_{2}=\quat r_{2}^{*}\quat r_{1}^{*}\quat p^{0}\quat r_{1}\quat r_{2}$,
which implies that the unit quaternion that represent the rotation
from $\frame 0$ to $\frame 2$ is given by $\quat r_{1}\quat r_{2}$,
showing again that a sequence of rotations is given by multiplication
of unit quaternions.

\begin{figure}[tbh]
\noindent \begin{centering}
\includegraphics[width=1\columnwidth]{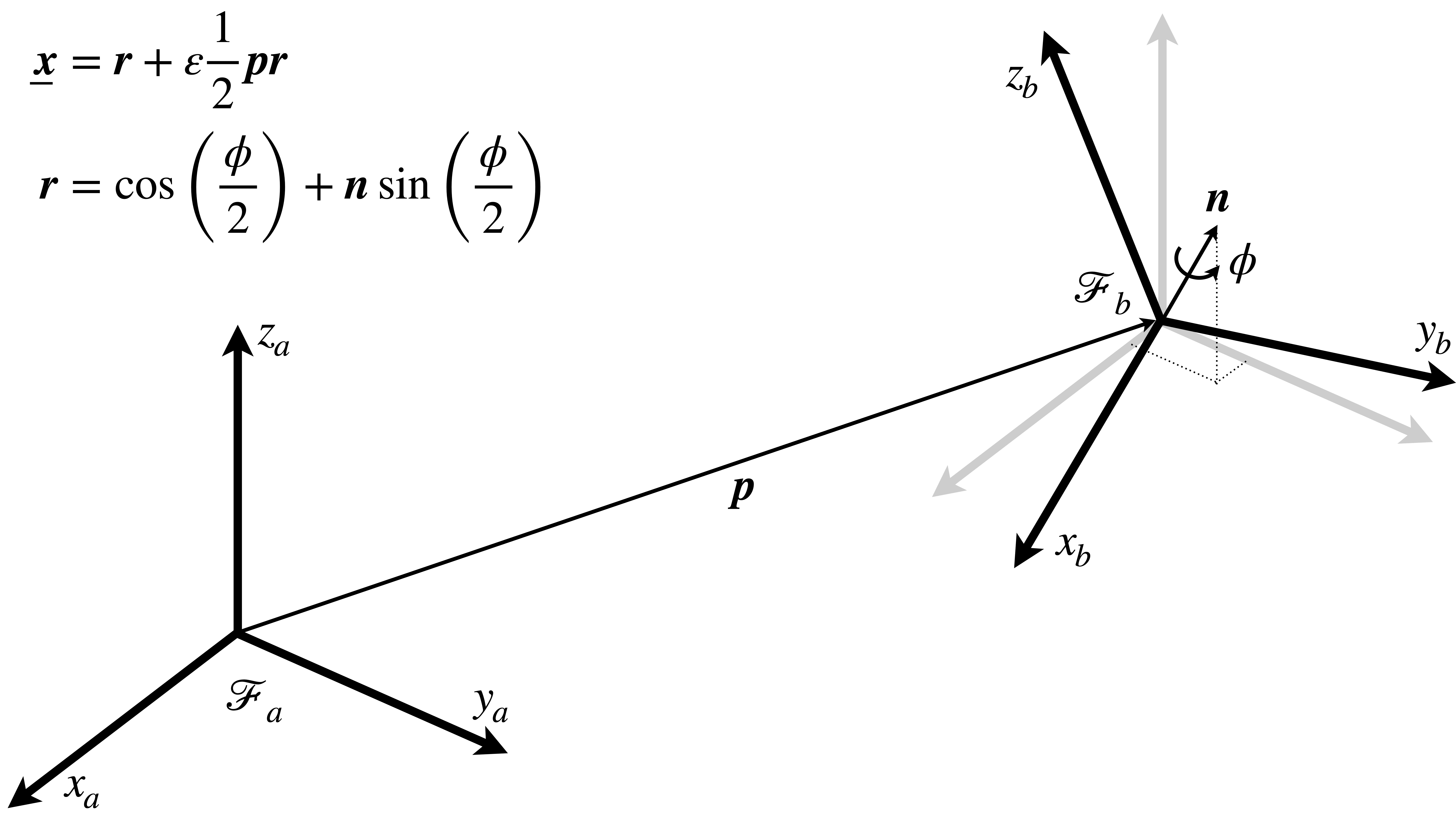}
\par\end{centering}
\caption{A rigid motion $\protect\dq x$ between frames $\protect\frame a$
and $\protect\frame b$ given by a translation $\protect\quat p$
followed by a rotation $\protect\quat r$.\label{fig:A-rigid-motion}}
\end{figure}

Translations and rotations can be grouped into a single unit dual
quaternion. More specifically, a rigid motion is represented by 
\begin{align}
\dq x & =\quat r+\dual\frac{1}{2}\quat p\quat r,\label{eq:unit_dual_quaternion}
\end{align}
as illustrated in Fig.~\ref{fig:A-rigid-motion}, and a sequence
of rigid motions is given by multiplication of unit dual quaternions.
For a more detailed description of the fundamentals of dual quaternion
algebra, see \cite{Adorno2017}.

Using DQ Robotics, a translation of $\left(0.1,0.2,0.3\right)\unit{m}$
followed by a rotation of $\pi$ rad around the $y$-axis, using \eqref{eq:unit_dual_quaternion},
is declared in MATLAB as

\begin{lstlisting}[caption={Declaring a unit dual quaternion in MATLAB.},label={lis:desired-end-effector-pose}]
r = cos(pi/2) + j_*sin(pi/2);
p = 0.1*i_ + 0.2*j_ + 0.3*k_;
xd = r + E_*0.5*p*r;
\end{lstlisting}
Alternatively, one could declare the rotation and translation as

\begin{lstlisting}
r = DQ([cos(pi/2), 0, sin(pi/2), 0]);
p = DQ([0, 0.1, 0.2, 0.3]);
\end{lstlisting}
Whereas the first way is closer to the mathematical notation, the
second one can be convenient when using the matrix functions available
in MATLAB, Python, and C++.

\section{Robot kinematics\label{sec:Robot-kinematics}}

\begin{figure}[tbh]
\noindent \begin{centering}
\includegraphics[width=1\columnwidth]{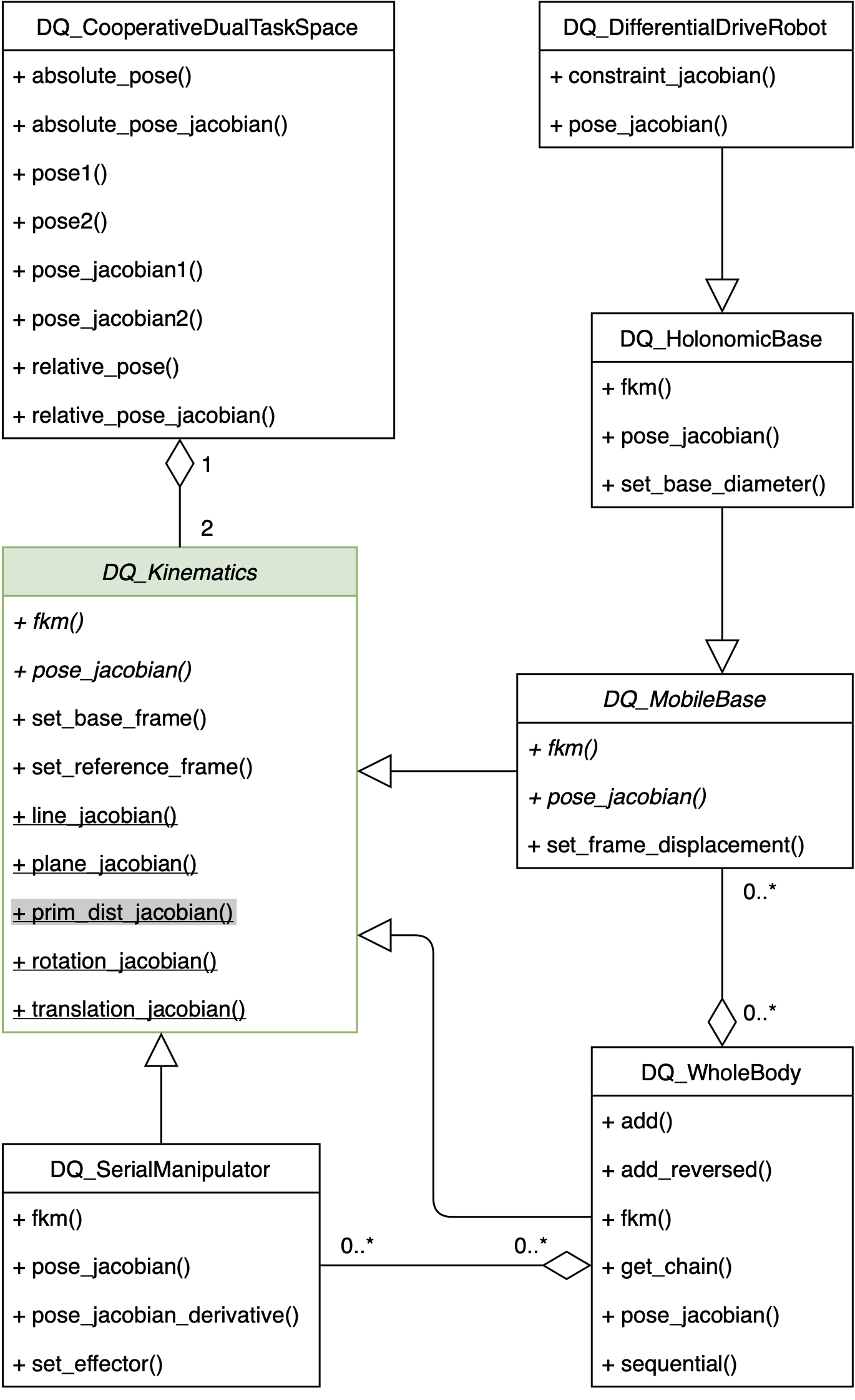}
\par\end{centering}
\caption{Simplified UML class diagram of the robot modeling classes with the
main methods that are available in C++, Python and MATLAB.\label{fig:Simplified-UML-diagram}}
\end{figure}

The implementation of robot kinematic modeling is done by means of
the abstract class \texttt{DQ\_Kinematics} and all its subclasses,
whose hierarchy is shown in Fig.~\ref{fig:Simplified-UML-diagram}.
The library currently supports serial manipulators, mobile bases and
their composition, which results in mobile manipulators (managed by
the \texttt{DQ\_WholeBody} class), bimanual systems (managed by the
\texttt{DQ\_CooperativeDualTaskSpace} class) and even branched mechanisms
such as bimanual mobile manipulators. Common manipulator robots are
available, such as the KUKA LWR4 and the Barrett WAM Arm, but custom
robots can be easily created by using arbitrary Denavit-Hartenberg
parameters. Some common mobile robots, such as the differential-drive
iRobot Create, are also available. Creating new ones requires only
the wheels radiuses and the axis length.

Let us consider the KUKA YouBot, which is a mobile manipulator composed
of a 5-DOF manipulator serially coupled to a holonomic base. One easy
way of defining it on DQ Robotics is to model the arm and the mobile
base separately and then assemble them to form a mobile manipulator.
The serial arm is defined, with the help of the Denavit-Hartenberg
parameters, as follows:
\begin{lstlisting}
arm_DH_theta = [0, pi/2, 0, pi/2, 0]; 
arm_DH_d = [0.147, 0, 0, 0, 0.218];             
arm_DH_a = [0, 0.155, 0.135, 0, 0];     
arm_DH_alpha = [pi/2, 0, 0, pi/2, 0];  
arm_DH_matrix = [arm_DH_theta; arm_DH_d; arm_DH_a;
                 arm_DH_alpha];
arm = DQ_SerialManipulator(arm_DH_matrix,'standard');
\end{lstlisting}
The holonomic base is defined as
\begin{lstlisting}
base = DQ_HolonomicBase();
\end{lstlisting}
and then they are coupled together in a \texttt{DQ\_WholeBody} object:
\begin{lstlisting}
x_bm = 1 + E_*0.5*(0.22575*i_ + 0.1441*k_);
base.set_frame_displacement(x_bm);
robot = DQ_WholeBody(base);
robot.add(arm);
\end{lstlisting}
Since there is a displacement between the mobile base frame and the
location where the arm is attached, we use the method \texttt{set\_frame\_displacement()}.
New kinematic chains can be serially coupled to the last chain by
using the method \texttt{add()}.

Because the KUKA Youbot is already defined in DQ Robotics, it suffices
to instantiate an object of the KukaYoubot class:
\begin{lstlisting}
youbot = KukaYoubot.kinematics();
\end{lstlisting}
All standard robots in DQ Robotics are defined inside the folder \texttt{{[}root\_folder{]}/robots}
and have the static method \texttt{kinematics() }that returns a \texttt{DQ\_Kinematics}
object. Therefore, a KUKA LWR 4 robot manipulator can be defined analogously:

\begin{lstlisting}
lwr4 = KukaLwr4Robot.kinematics(); 
\end{lstlisting}

All \texttt{DQ\_Kinematics} subclasses have common functions for robot
kinematics such as the ones used to calculate the forward kinematics,
the Jacobian matrix that maps the configuration velocities to the
time-derivative of the end-effector pose, as well as other Jacobian
matrices that map the configuration velocities to the time derivative
of other geometrical primitives attached to the end-effector, such
as lines and planes. This includes \texttt{DQ\_WholeBody} objects,
therefore the corresponding Jacobians are whole-body Jacobians that
take into account the complete kinematic chain. The list of the main
methods is summarized in the simplified UML diagram in Fig.~\ref{fig:Simplified-UML-diagram}
and detailed in Table~\ref{tab:DQ_Kinematics-and-subclasses}.

\noindent 
\begin{table*}[t]
\caption{Main methods of class \texttt{DQ\_Kinematics} and its subclasses.\label{tab:DQ_Kinematics-and-subclasses}}

\renewcommand*\arraystretch{1}
\noindent \begin{centering}
\begin{tabular*}{1\textwidth}{@{\extracolsep{\fill}}>{\raggedright}p{0.2\textwidth}>{\raggedright}p{0.75\textwidth}}
\toprule 
\multicolumn{2}{c}{\texttt{\textbf{\emph{DQ\_Kinematics}}}}\tabularnewline
\texttt{\textit{\scriptsize{}fkm}} & Compute the forward kinematics.\tabularnewline
\texttt{\textit{\scriptsize{}pose\_jacobian}} & Compute the Jacobian that maps the configurations velocities to the
time derivative of the pose of a frame attached to the robot.\tabularnewline
\texttt{\textit{\emph{\scriptsize{}set\_base\_frame}}} & Set the physical location of the robot in space.\tabularnewline
\texttt{\textit{\emph{\scriptsize{}set\_reference\_frame}}} & Set the reference frame for all calculations.\tabularnewline
\texttt{\textit{\emph{\scriptsize{}\uline{line\_jacobian}}}} & Compute the Jacobian that maps the configurations velocities to the
time derivative of a line attached to the robot.\tabularnewline
\texttt{\textit{\emph{\scriptsize{}\uline{plane\_jacobian}}}} & Compute the Jacobian that maps the configurations velocities to the
time derivative of a plane attached to the robot.\tabularnewline
\texttt{\textit{\emph{\scriptsize{}\uline{rotation\_jacobian}}}} & Compute only the rotational part of the \texttt{pose\_jacobian}.\tabularnewline
\texttt{\textit{\emph{\scriptsize{}\uline{translation\_jacobian}}}} & Compute only the translational part of the \texttt{pose\_jacobian}.\tabularnewline
\texttt{\textit{\emph{\scriptsize{}\uline{\{prim\_dist\_jacobian\}}}}} & Compute the Jacobian that maps the configurations velocities to the
time derivative of the distance between a primitive attached to the
robot and another primitive in the workspace. For instance, \texttt{\{prim\_dist\_jacobian\}}
can be \texttt{line\_to\_line\_distance\_jacobian}, \texttt{line\_to\_point\_distance\_jacobian},
\texttt{plane\_to\_point\_distance\_jacobian}, \texttt{point\_to\_line\_distance\_jacobian},
\texttt{point\_to\_plane\_distance\_jacobian}, \texttt{point\_to\_point\_distance\_jacobian}.\tabularnewline
\midrule 
\multicolumn{2}{c}{\texttt{\textbf{DQ\_SerialManipulator}}}\tabularnewline
\texttt{\textit{\emph{\scriptsize{}pose\_jacobian\_derivative}}} & Compute the analytical time derivative of the pose Jacobian matrix.\tabularnewline
\texttt{\textit{\emph{\scriptsize{}set\_effector}}} & Set a constant transformation for the end-effector pose with respect
to the frame attached to the end of the last link.\tabularnewline
\midrule 
\multicolumn{2}{c}{\texttt{\textbf{\emph{DQ\_MobileBase}}}, \texttt{\textbf{DQ\_HolonomicBase}},
and \texttt{\textbf{DQ\_DifferentialDriveRobot}}}\tabularnewline
\texttt{\textit{\emph{\scriptsize{}set\_frame\_displacement}}} & Set the rigid transformation for the base frame.\tabularnewline
\texttt{\textit{\emph{\scriptsize{}set\_base\_diameter}}} & Change the base diameter.\tabularnewline
\texttt{\textit{\emph{\scriptsize{}constraint\_jacobian}}} & Compute the Jacobian that relates the wheels velocities to the configuration
velocities.\tabularnewline
\midrule 
\multicolumn{2}{c}{\texttt{\textbf{DQ\_WholeBody}}}\tabularnewline
\texttt{\textit{\emph{\scriptsize{}add}}} & Add a new element to the end of the serially coupled kinematic chain.\tabularnewline
\texttt{\textit{\emph{\scriptsize{}add\_reversed}}} & Add a new element, but in reverse order, to the end of the serially
coupled kinematic chain.\tabularnewline
\texttt{\textit{\emph{\scriptsize{}get\_chain}}} & Returns the complete kinematic chain.\tabularnewline
\texttt{\textit{\emph{\scriptsize{}sequential}}} & Reorganize a sequential configuration vector in the ordering required
by each kinematic chain (i.e., the vector blocks corresponding to
reversed chains are reversed).\tabularnewline
\midrule
\multicolumn{2}{c}{\texttt{\textbf{DQ\_CooperativeDualTaskSpace}}}\tabularnewline
\texttt{\scriptsize{}absolute\_pose} & Compute the pose of a frame between the two end-effectors, which is
usually related to a grasped object.\tabularnewline
\texttt{\scriptsize{}absolute\_pose\_jacobian} & Compute the Jacobian that maps the joint velocities of the two-arm
system to the time derivative of the absolute pose.\tabularnewline
\texttt{\scriptsize{}pose1 }and\texttt{\scriptsize{} pose2} & Compute the poses of the first and second end-effectors, respectively.\tabularnewline
\texttt{\scriptsize{}pose\_jacobian1 }and\texttt{\scriptsize{} pose\_jacobian2} & Compute the Jacobians that maps the configurations velocities to the
time derivative of the poses of the first and second end-effectors,
respectively.\tabularnewline
\texttt{\scriptsize{}relative\_pose} & Compute the rigid transformation between the two end-effectors.\tabularnewline
\texttt{\scriptsize{}relative\_pose\_jacobian} & Compute the Jacobian that maps the joint velocities of the two-arm
system to the time derivative of the relative pose.\tabularnewline
\bottomrule
\end{tabular*}
\par\end{centering}
\medskip{}

{*}Abstract methods are written in \textit{italics}\emph{,} concrete
methods are written in upright, and static methods are \uline{underlined}.
Concrete methods that implement their abstract counterparts are omitted
for the sake of conciseness.
\end{table*}

\section{Geometric Primitives\label{sec:Geometric-Primitives}}

\begin{figure}[tbh]
\noindent \begin{centering}
\includegraphics[width=1\columnwidth]{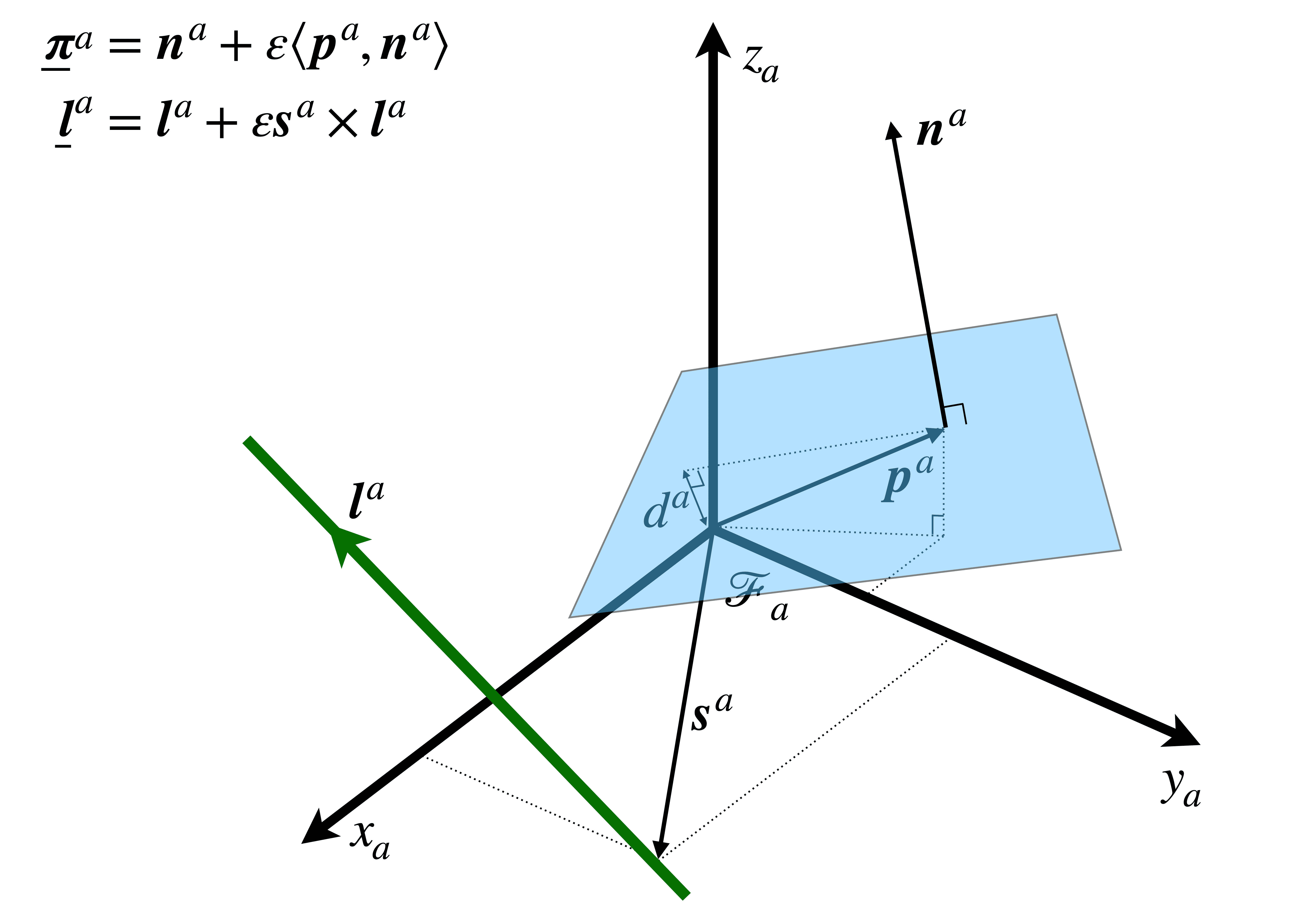}
\par\end{centering}
\caption{Two geometric primitives represented by elements of the dual quaternion
algebra: (a) the \emph{blue} plane $\protect\dq{\pi}^{a}$, represented
by the normal $\protect\quat n^{a}$ and the distance to the origin
$d^{a}=\protect\dotproduct{\protect\quat p^{a},\protect\quat n^{a}}$,
where $\protect\quat p^{a}$ is an arbitrary point on the plane; (b)
and the \emph{green }line $\protect\dq l^{a}$, represented by the
direction $\protect\quat l^{a}$ and the moment $\protect\quat s^{a}\times\protect\quat l^{a}$,
where $\protect\quat s^{a}$ is an arbitrary point on the line.\label{fig:plane_line}}
\end{figure}

Several geometrical primitives, such as points, planes, and lines
are represented as elements of the dual quaternion algebra \cite{Adorno2017},
which is particularly useful when incorporating geometric constraints
into robot motion controllers. For instance, a plane expressed in
a frame $\frame a$ is given by $\dq{\pi}^{a}=\quat n^{a}+\dual d^{a}$,
where $\quat n^{a}=n_{x}\imi+n_{y}\imj+n_{z}\imk$ is the normal to
the plane and $d^{a}=\dotproduct{\quat p^{a},\quat n^{a}}$ is the
signed distance between the plane and the origin of $\frame a$, and
$\quat p^{a}$ is an arbitrary point on the plane, as illustrated
in Fig.~\ref{fig:plane_line}. 

Therefore, the plane $\dq{\pi}^{a}=\imj+\dual1.5$ (i.e., a plane
whose normal is parallel to the $y$-axis and whose distance from
$\frame a$ is 1.5 m) is defined in DQ Robotics as

\begin{figure}[tbh]
\noindent \begin{centering}
\includegraphics[width=1\columnwidth]{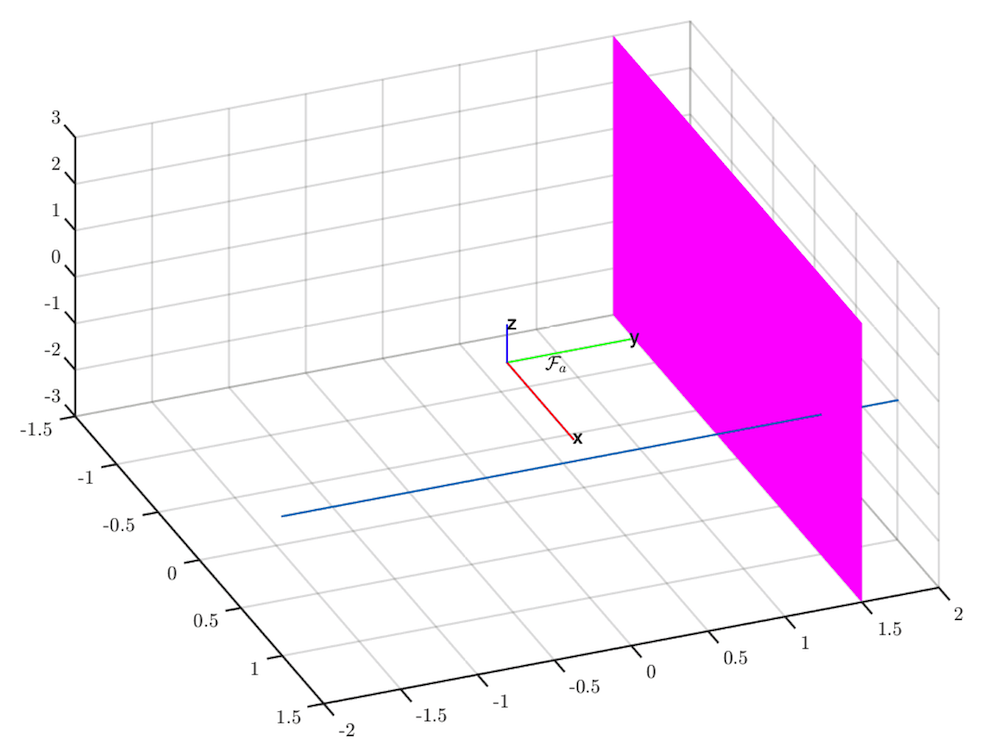}
\par\end{centering}
\caption{MATLAB plot: reference frame $\protect\frame a$; line $\protect\dq l^{a}=\protect\imj+\protect\dual(\protect\imi\times\protect\imj)$
parallel to the $y$-axis and passing through point $(1,0,0)$; and
plane $\protect\dq{\pi}^{a}=\protect\imj+\protect\dual1.5$ with normal
parallel to the $y$-axis and whose distance from the origin of $\protect\frame a$
is 1.5 m.\label{fig:matlab_plot}}
\end{figure}

\begin{lstlisting}
plane_a = j_ + E_ * 1.5;
\end{lstlisting}
Analogously, a line in $\frame a$, with direction given by $\quat l^{a}=l_{x}\imi+l_{y}\imj+l_{z}\imk$
and passing through point $\quat s^{a}=p_{x}\imi+p_{y}\imj+p_{z}\imk$,
is represented in dual quaternion algebra as $\dq l^{a}=\quat l^{a}+\dual\quat s^{a}\times\quat l^{a}$,
as illustrated in Fig.~\ref{fig:plane_line}. Therefore, a line parallel
to the $y$-axis passing through point $\quat s^{a}=\imi$ (i.e.,
with coordinates $\left(1,0,0\right)$) is defined in DQ Robotics
as
\begin{lstlisting}
line_a = j_ + E_ * cross(i_,j_);
\end{lstlisting}
In MATLAB, the plot functions are available for frames, planes, and
lines:

\begin{lstlisting}
plot(DQ(1),'name', '$\mathcal{F}_a$');
plot(plane_a, 'plane', 20, 'color', 'magenta');
plot(line_a, 'line', 5);
\end{lstlisting}
The result is shown in Fig.~\ref{fig:matlab_plot}.

The library implements the utility class \texttt{DQ\_Geometry}, which
provides methods for geometric calculations between primitives, such
as distances. Jacobians relating the robot joint velocities to primitives
attached to the end-effector are available in the \texttt{DQ\_Kinematics}
class, such as \texttt{line\_jacobian()} and \texttt{plane\_jacobian()}.
In addition, \texttt{DQ\_Kinematics} contains several Jacobians that
relate the time derivative of the distance between two primitives
(e.g., \texttt{plane\_to\_point\_distance\_jacobian()}) which are
grouped in the UML class diagram of Fig.~\ref{fig:Simplified-UML-diagram}
as \texttt{prim\_dist\_jacobian()}.

\section{Robot motion control}

Several advantages make dual quaternion algebra attractive when designing
controllers. On the one hand, unit dual quaternions are easily mapped
into a vector structure. The vector structure is particularly convenient
in pose control as there is no need to use intermediate mappings nor
extract parameters from the dual quaternion. On the other hand, the
position is easily extracted from the HTM, but to obtain the orientation
one usually has to extract the rotation angle and the rotation axis
from the rotation matrix. This extraction introduces representational
singularities when the rotation angle equals zero. Furthermore, when
designing constrained controllers based on geometrical constraints,
one has to represent several geometrical primitives in different coordinate
systems and in different locations. Dual quaternion algebra is particularly
useful in this case because several relevant geometrical primitives
are easily represented as dual quaternions, as shown in Section~\ref{sec:Geometric-Primitives}.

Users can implement their own controllers in DQ Robotics. For instance,
suppose that the desired end-effector pose is given by Listing~\ref{lis:desired-end-effector-pose}
and the initial robot configuration and controller parameters are
given in Listing~\ref{lis:parameters-simple-kinematic-controller}.

\begin{lstlisting}[caption={Parameters for a simple kinematic controller.},label={lis:parameters-simple-kinematic-controller}]
q = [0, 0.3770, 0.1257, -0.5655, 0, 0, 0]'; 
T = 0.001;     % sampling time 
gain = 10;     % controller gain
\end{lstlisting}
A classic controller based on an Euclidean error function and the
Jacobian pseudo-inverse can be implemented in MATLAB as shown in Listing~\ref{lis:Kinematic-controller-based-jacobian-pseudoinverse}.
\begin{lstlisting}[caption={Kinematic controller based on the Jacobian pseudoinverse.},label={lis:Kinematic-controller-based-jacobian-pseudoinverse},tabsize=4]
e = ones(8,1); % initialize the error vector 
while norm(e) > 0.001     
	J = lwr4.pose_jacobian(q);      
	x = lwr4.fkm(q);     
	e = vec8(x-xd);     
	u = -pinv(J)*gain*e;     
	q = q + T*u;  
end
\end{lstlisting}
In Listing~\ref{lis:Kinematic-controller-based-jacobian-pseudoinverse},
when executing the controller on a real robot, the control input \texttt{u}
is usually sent directly to the robot.

\begin{figure}[tbh]
\noindent \begin{centering}
\includegraphics[width=1\columnwidth]{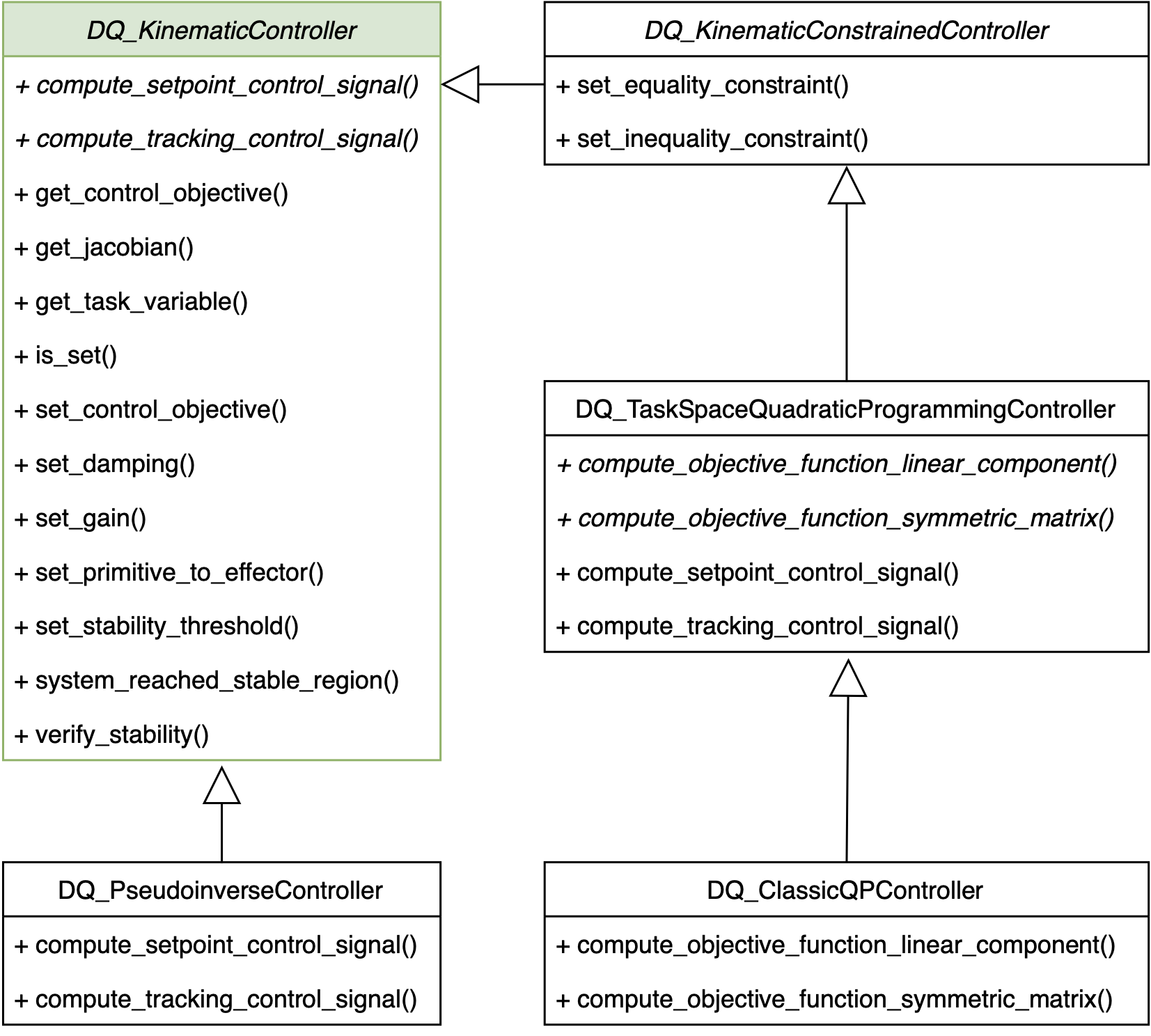}
\par\end{centering}
\caption{Simplified UML class diagram of the robot kinematic controller classes
with the main methods that are available in C++, Python and MATLAB.\label{fig:UML-Diagram-Kinematic-Controller-Classes}}
\end{figure}

There are several kinematic controllers included in DQ Robotics, including
the one in Listing~\ref{lis:Kinematic-controller-based-jacobian-pseudoinverse}.
The class hierarchy for the kinematic controllers is summarized by
the simplified UML diagram in Fig.~\ref{fig:UML-Diagram-Kinematic-Controller-Classes}.
All kinematic controller classes inherit from the abstract superclass
\texttt{DQ\_KinematicController}, and their main methods are detailed
in Table~\ref{tab:DQ_KinematicController-methods}.

\noindent 
\begin{table*}[tbh]
\caption{Main methods of class \texttt{DQ\_KinematicController} and its main
subclasses.\label{tab:DQ_KinematicController-methods}}

\renewcommand*\arraystretch{1}
\noindent \begin{centering}
\begin{tabular*}{1\textwidth}{@{\extracolsep{\fill}}>{\raggedright}p{0.35\textwidth}>{\raggedright}p{0.6\textwidth}}
\toprule 
\multicolumn{2}{c}{\texttt{\textbf{\emph{DQ\_KinematicController}}}}\tabularnewline
\texttt{\textit{\scriptsize{}compute\_setpoint\_control\_signal}} & Compute the control input to regulate to a set-point.\tabularnewline
\texttt{\textit{\scriptsize{}compute\_tracking\_control\_signal}} & Compute the control input to track a trajectory.\tabularnewline
\texttt{\textit{\emph{\scriptsize{}get\_control\_objective}}} & \textit{\emph{Return the control objective.}}\tabularnewline
\texttt{\textit{\emph{\scriptsize{}get\_jacobian}}} & \textit{\emph{Return the correct Jacobian based on the control objective.}}\tabularnewline
\texttt{\textit{\emph{\scriptsize{}get\_task\_variable}}} & \textit{\emph{Return the task variable based on the control objective.}}\tabularnewline
\texttt{\textit{\emph{\scriptsize{}is\_set}}} & \textit{\emph{Verify if the controller is set and ready to be used.}}\tabularnewline
\texttt{\textit{\emph{\scriptsize{}set\_control\_objective}}} & \textit{\emph{Set the control objective using predefined goals in
ControlObjective.}}\tabularnewline
\texttt{\textit{\emph{\scriptsize{}set\_damping}}} & Set the damping to prevent instabilities near singular configurations.\tabularnewline
\texttt{\textit{\emph{\scriptsize{}set\_gain}}} & \textit{\emph{Set the controller gain.}}\tabularnewline
\texttt{\textit{\emph{\scriptsize{}set\_primitive\_to\_effector}}} & \textit{\emph{Attach primitive (e.g., plane, line, point) to the end-effector.}}\tabularnewline
\texttt{\textit{\emph{\scriptsize{}set\_stability\_threshold}}} & \textit{\emph{Set the threshold that determines if a stable region
has been reached.}}\tabularnewline
\texttt{\textit{\emph{\scriptsize{}system\_reached\_stable\_region}}} & Return \texttt{true} if the trajectories of the closed-loop system
have reached a stable region (i.e., a positive invariant set), \texttt{false}
otherwise.\tabularnewline
\texttt{\textit{\emph{\scriptsize{}verify\_stability}}} & \textit{\emph{Verify if the closed-loop region has reached a stable
region.}}\tabularnewline
\midrule 
\multicolumn{2}{c}{\texttt{\textbf{DQ\_PseudoinverseSetpointController}}}\tabularnewline
\texttt{\scriptsize{}compute\_tracking\_control\_signal} & Given the task error $\tilde{\myvec x}=\myvec x-\myvec x_{d}\in\mathbb{R}^{n}$,
and the feedforward term $\dot{\myvec x}_{d}$, compute the control
signal $\myvec u=\mymatrix J^{+}\left(-\lambda\tilde{\myvec x}+\dot{\myvec x}_{d}\right)$,
where $\mymatrix J^{+}$ is the pseudoinverse of the task Jacobian
$\mymatrix J$ and $\lambda$ is the controller gain.\tabularnewline
\midrule 
\multicolumn{2}{c}{\texttt{\textbf{\emph{DQ\_KinematicConstrainedController}}}}\tabularnewline
\texttt{\scriptsize{}set\_equality\_constraint} & Add equality constraints of type $\mymatrix B\dot{\myvec q}=\myvec b$,
where $\dot{\myvec q}\in\mathbb{R}^{n}$ is the vector of joint velocities,
$\myvec b\in\mathbb{R}^{m}$ is the vector of equality constraints
and $\mymatrix B\in\mathbb{R}^{m\times n}$.\tabularnewline
\texttt{\scriptsize{}set\_inequality\_constraint} & Add inequality constraints of type $\myvec A\dot{\myvec q}\preceq\myvec a$,
where $\dot{\myvec q}\in\mathbb{R}^{n}$ is the vector of joint velocities,
$\myvec a\in\mathbb{R}^{m}$ is the vector of inequality constraints
and $\mymatrix A\in\mathbb{R}^{m\times n}$.\tabularnewline
\midrule 
\multicolumn{2}{c}{\texttt{\textbf{\emph{DQ\_TaskspaceQuadraticProgrammingController}}}}\tabularnewline
\texttt{\emph{\scriptsize{}compute\_objective\_function\_linear\_component}} & Compute the vector $\myvec h$ used in the objective function $(1/2)\dot{\myvec q}^{T}\mymatrix H\dot{\myvec q}+\myvec h^{T}\dot{\myvec q}$.\tabularnewline
\texttt{\emph{\scriptsize{}compute\_objective\_function\_symmetric\_matrix}} & Compute the matrix $\mymatrix H$ used in the objective function $(1/2)\dot{\myvec q}^{T}\mymatrix H\dot{\myvec q}+\myvec h^{T}\dot{\myvec q}$.\tabularnewline
\texttt{\scriptsize{}compute\_tracking\_control\_signal} & Given the task error $\tilde{\myvec x}=\myvec x-\myvec x_{d}\in\mathbb{R}^{n}$,
compute the control signal given by $\myvec u\in\arg\min_{\dot{\myvec q}}(1/2)\dot{\myvec q}^{T}\mymatrix H\dot{\myvec q}+\myvec h^{T}\dot{\myvec q}$
subject to $\mymatrix A\dot{\myvec q}\preceq\myvec a$ and $\mymatrix B\dot{\myvec q}=\myvec b$.\tabularnewline
\midrule
\multicolumn{2}{c}{\texttt{\textbf{DQ\_ClassicQPController}}}\tabularnewline
\texttt{\scriptsize{}compute\_objective\_function\_linear\_component} & Compute the vector $\myvec h$ used in $(1/2)\dot{\myvec q}^{T}\mymatrix H\dot{\myvec q}+\myvec h^{T}\dot{\myvec q}=\norm{\mymatrix J\dot{\myvec q}+\lambda\tilde{\myvec x}-\dot{\myvec x}_{d}}_{2}^{2}$.\tabularnewline
\texttt{\scriptsize{}compute\_objective\_function\_symmetric\_matrix} & Compute the matrix $\mymatrix H$ used in $(1/2)\dot{\myvec q}^{T}\mymatrix H\dot{\myvec q}+\myvec h^{T}\dot{\myvec q}=\norm{\mymatrix J\dot{\myvec q}+\lambda\tilde{\myvec x}-\dot{\myvec x}_{d}}_{2}^{2}$.\tabularnewline
\bottomrule
\end{tabular*}
\par\end{centering}
\medskip{}

{*}Abstract methods are written in \textit{italics}\emph{ }and\emph{
}concrete methods are written in upright. In all concrete classes,
the method \texttt{compute\_setpoint\_control\_signal()} does not
include the feedforward term in the method parameters and is equivalent
to the method \texttt{compute\_tracking\_control\_signal} with $\dot{\myvec x}_{d}=\myvec 0$.
Therefore, they are omitted for the sake of conciseness.
\end{table*}

The \texttt{DQ\_KinematicController} subclasses are focused on the
control structure, not on particular geometrical tasks. The same subclass
can be used regardless if the goal is to control the end-effector
pose, position, orientation, etc. In order to distinguish between
task objectives, it suffices to use
\begin{lstlisting}
controller.set_control_objective(GOAL)
\end{lstlisting}
where \texttt{GOAL} is an object from the enumeration class \texttt{ControlObjective},
which currently provides the following enumeration members: \texttt{Distance},
\texttt{DistanceToPlane},\texttt{ Line}, \texttt{Plane}, \texttt{Pose},
\texttt{Rotation}, and \texttt{Translation}. Therefore, the implementation
in Listing~\ref{lis:Kinematic-controller-based-jacobian-pseudoinverse}
can be rewritten using the \texttt{DQ\_PseudoinverseController} class
as 
\begin{lstlisting}[caption={Controller based on the \texttt{DQ\_PseudoinverseController} class.},label={lis:Controller-based-on-DQ_PseudoinverseController},basicstyle={\scriptsize\ttfamily}]
control = DQ_PseudoinverseController(lwr4);
control.set_control_objective(ControlObjective.Pose);

control.set_gain(gain);
control.set_stability_threshold(0.0001);
while ~control.system_reached_stable_region()          
    u = control.compute_setpoint_control_signal(q,vec8(xd));
    q = q + T*u; 
end        
\end{lstlisting}

Furthermore, if the geometric task objective changes, the \texttt{DQ\_KinematicController}
subclass is responsible for calculating the appropriate forward kinematics
and the corresponding Jacobian. For instance, if a line is attached
to the end-effector and the goal is to align it with a line in the
workspace, then it suffices to replace the second line of Listing~\ref{lis:Controller-based-on-DQ_PseudoinverseController}
by
\begin{lstlisting}
control.set_primitive_to_effector(line);
control.set_control_objective(ControlObjective.Line);
\end{lstlisting}
where \texttt{line} is a dual quaternion corresponding to a line passing
through the origin of the end-effector frame.

Furthermore, since any subclass of \texttt{DQ\_KinematicController}
provides the methods used in Listing~\ref{lis:Controller-based-on-DQ_PseudoinverseController},
changing the controller is a matter of changing just one line of code,
namely the first one in Listing~\ref{lis:Controller-based-on-DQ_PseudoinverseController}.

\section{Interface with ROS}

The Robot Operating System (ROS) \cite{quigley2009ros} is widely
used by the robotics community. After the one-line installation described
in Section~\ref{sec:development_infrastructure}, the Python version
of the DQ Robotics library can be imported in any Python script in
the system; therefore, it is compatible with ROS out-of-the-box. MATLAB
also has an interface with ROS.\footnote{https://www.mathworks.com/help/ros/ug/get-started-with-ros.html}

For C++ code, in its most recent versions, ROS uses the catkin\_build\footnote{https://catkin-tools.readthedocs.io/en/latest/verbs/catkin\_build.html}
environment, which itself depends on CMAKE. Given that DQ Robotics
C++ is installed as a system library in Ubuntu, as shown in Section~\ref{sec:development_infrastructure},
ROS users can readily have access to the library as they would to
any other system library with a proper CMAKE configuration. After
installation, the C++ version of the library can be linked with

\begin{lstlisting}[language=sh]
target_link_libraries(my_binary dqrobotics) 
\end{lstlisting}
for a given binary called my\_binary.

\section{Interface with V-REP}

DQ Robotics provides a simple interface to V-REP \cite{Rohmer2013},
enabling users to develop complex simulations without having to delve
into the V-REP documentation. The V-REP interface comes bundled with
the MATLAB and Python versions of the library. The C++ version of
our V-REP interface comes as a separate package and the following
CMAKE directive will link the required shared object

\begin{lstlisting}[language=sh]
target_link_libraries(my_binary 
                      dqrobotics 
                      dqrobotics-vrep-interface) 
\end{lstlisting}
for a given my\_binary.

The available methods are shown in Table~\ref{tab:VrepInterface}.

\noindent 
\begin{table}[tbh]
\caption{\texttt{VrepInterface}.\label{tab:VrepInterface}}

\noindent \begin{centering}
\begin{tabular*}{1\columnwidth}{@{\extracolsep{\fill}}>{\raggedright}p{0.35\columnwidth}>{\raggedright}p{0.6\columnwidth}}
\toprule 
\texttt{\scriptsize{}connect} & Connect to a V-REP Remote API Server.\tabularnewline
\texttt{\scriptsize{}disconnect} & Disconnect from currently connected server.\tabularnewline
\texttt{\scriptsize{}disconnect\_all} & Flush all Remote API connections.\tabularnewline
\texttt{\scriptsize{}start\_simulation} & Start V-REP simulation.\tabularnewline
\texttt{\scriptsize{}stop\_simulation} & Stop V-REP simulation.\tabularnewline
\texttt{\scriptsize{}get\_object\_translation} & Get object translation as a pure quaternion.\tabularnewline
\texttt{\scriptsize{}set\_object\_translation} & Set object translation with a pure quaternion.\tabularnewline
\texttt{\scriptsize{}get\_object\_rotation} & Get object rotation as a unit quaternion.\tabularnewline
\texttt{\scriptsize{}set\_object\_rotation} & Set object rotation with a unit quaternion.\tabularnewline
\texttt{\scriptsize{}get\_object\_pose} & Get object pose as a unit dual quaternion.\tabularnewline
\texttt{\scriptsize{}set\_object\_pose} & Set object pose with a unit dual quaternion.\tabularnewline
\texttt{\scriptsize{}set\_joint\_positions} & Set the joint positions of a manipulator robot.\tabularnewline
\texttt{\scriptsize{}set\_joint\_target\_positions} & Set the joint target positions of a manipulator robot.\tabularnewline
\texttt{\scriptsize{}get\_joint\_positions} & Get the joint positions of a manipulator robot.\tabularnewline
\bottomrule
\end{tabular*}
\par\end{centering}
\medskip{}
\end{table}

To start the communication with V-REP using its remote API,\footnote{\url{http://www.coppeliarobotics.com/helpFiles/en/legacyRemoteApiOverview.htm}}
only two methods are necessary:

\begin{lstlisting}
vi = DQ_VrepInterface; 
vi.connect('127.0.0.1',19997); 
vi.start_simulation();
\end{lstlisting}
The method \texttt{start\_simulation} starts the V-REP simulation
with the default asynchronous mode and the recommended 5 ms communication
thread cycle.\footnote{For other parameters, refer to the code.}

Since each robot joint in V-REP is associated with a name, it is convenient
to encapsulate this information inside the robot class. Therefore,
classes implementing robots that use the V-REP interface must be a
subclass of \texttt{DQ\_VrepRobot} to provide the methods \texttt{send\_q\_to\_vrep},
and \texttt{get\_q\_from\_vrep}, which are used to send the desired
robot configuration vector to V-REP and get the current robot configuration
vector from V-REP, respectively.

Objects poses in a V-REP scene can be retrieved or set by using the
methods \texttt{get\_object\_pose} and \texttt{set\_object\_pose},
respectively, which are useful when designing motion planners or controllers
that take into account the constraints imposed by obstacles. Finally,
in order to end the V-REP simulation it suffices to use two methods:
\begin{lstlisting}
vi.stop_simulation(); 
vi.disconnect();
\end{lstlisting}

\subsection{A more complete example}

\begin{figure}[tbh]
\noindent \begin{centering}
\includegraphics[width=1\columnwidth]{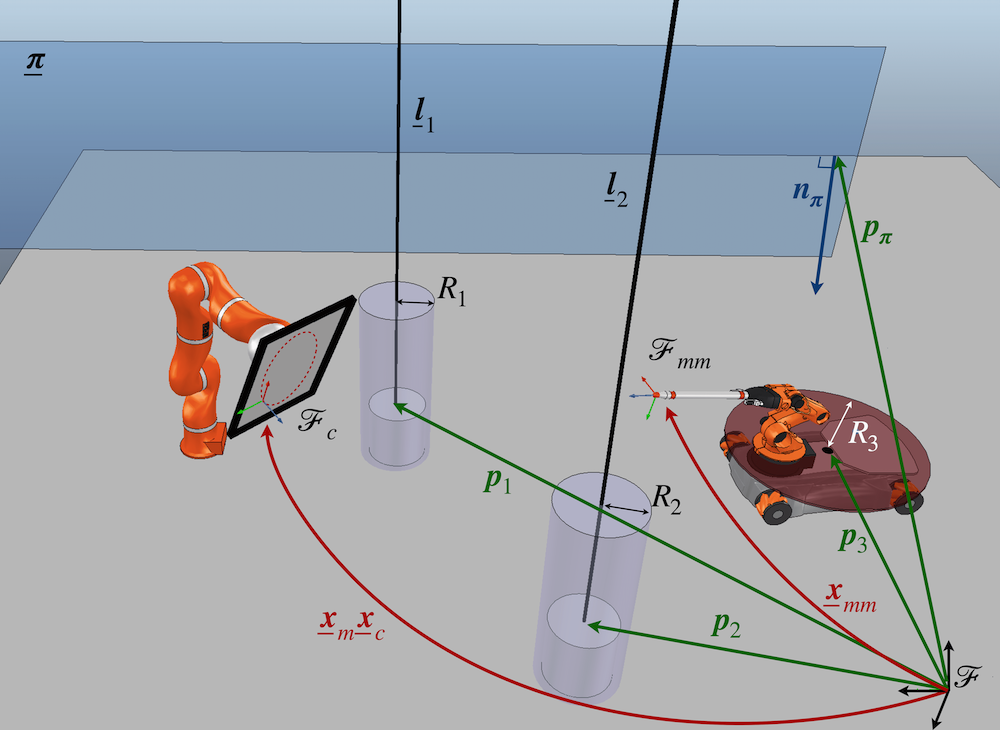}
\par\end{centering}
\caption{A more complete example. The reference frame is given by $\protect\frame{}$,
the time-varying frame $\protect\frame c$ is used to indicate where
the circle must be drawn in the whiteboard, and $\protect\frame{mm}$
is the end-effector frame of the mobile manipulator. Obstacles are
represented as geometrical primitives: the wall is represented by
the plane $\protect\dq{\pi}=\protect\quat n_{\protect\dq{\pi}}+\protect\dual\protect\dotproduct{\protect\quat p_{\protect\dq{\pi}},\protect\quat n_{\protect\dq{\pi}}}$,
where $\protect\quat p_{\protect\dq{\pi}}$ is an arbitrary point
on the plane and $\protect\quat n_{\protect\dq{\pi}}$ is the plane
normal, and the two cylindrical obstacles are represented by $\left(\protect\dq l_{1},R_{1}\right)$
and $\left(\protect\dq l_{2},R_{2}\right)$, where $\protect\dq l_{i}=\protect\quat l_{i}+\protect\dual\protect\quat p_{i}\times\protect\quat l_{i}$
is the ith cylinder centerline and $R_{i}$ is the corresponding cylinder
radius with $i\in\{1,2\}$. Furthermore, $\protect\quat l_{i}$ is
the line direction and $\protect\quat p_{i}$ is an arbitrary point
on the ith line. To prevent collisions between the robot and the obstacles,
the robot is represented by a circle given by $\left(\protect\quat p_{3},R_{3}\right)$,
where $\protect\quat p_{3}$ and $R_{3}$ are the center and radius
of the circle, respectively.\label{fig:A-more-complete-example}}
\end{figure}

To better highlight how DQ Robotics can be used with V-REP, in this
example, a KUKA LWR4 manipulator robot interacts with a KUKA YouBot
mobile manipulator in a workspace containing three obstacles, namely
two cylinders and a plane, as shown in Fig.~\ref{fig:A-more-complete-example}.

The manipulator robot holds a whiteboard and uses a pseudo-inverse-based
controller with a feedforward term to track a trajectory, given by
$\dq x_{m}(t)=\quat r_{m}(t)\dq x_{m}(0)\quat p_{m}(t)$, where $\quat r_{m}=\cos(\phi(t)/2)+\imk\sin(\phi(t)/2)$
with $\phi(t)=(\pi/2)\sin(\omega_{n}t)$, and $\quat p=1+(1/2)\dual d_{z}\cos(\omega_{d}t)\imk$.
Viewed from the top, the whiteboard will follow a semi-circular path
with a radial oscillation at a frequency of $\omega_{d}\,\unit{rad/s}$
and amplitude $d_{z}\,\unit{m}$, and an oscillation around the vertical
axis at a frequency of $\omega_{n}\,\unit{rad/s}$. The mobile manipulator
holds a felt pen and follows the manipulator trajectory while drawing
a circle on the whiteboard. The mobile manipulator desired end-effector
trajectory is given by $\dq x_{mm}(t)=\dq x_{m}(t)\dq x_{c}\imj$,
where $\dq x_{c}=1+(1/2)\dual0.015\imk$ is a constant displacement
of $1.5\unit{cm}$ along the $z$-axis to account for the whiteboard
width and $\imj=\cos(\pi/2)+\imj\sin(\pi/2)$ is a rotation of $\pi$
around the $y$-axis so that both end-effectors have their $z$-axis
pointing to opposite directions. To prevent collision with obstacles
it uses a constrained controller, in which for each obstacle we use
a differential inequality to ensure that the robot will approach it,
in the worst case, with an exponential velocity decrease. For each
obstacle, we define a distance metric ${\tilde{d}\left(t\right)\triangleq d\left(t\right)-d_{\mathrm{safe}}}$,
where $d_{\mathrm{safe}}\in\left[0,\infty\right)$ is an arbitrary
constant safe distance, and the following inequalities must hold for
all $t$ :
\begin{align}
\dot{\tilde{d}}\left(t\right)\geq- & \eta_{d}\tilde{d}\left(t\right)\Longleftrightarrow-\mymatrix J_{d}\dot{\myvec q}\leq\eta_{d}\tilde{d}\left(t\right),\label{eq:Constraint}
\end{align}
where $\eta_{d}\in[0,\infty)$ is used to adjust the approach velocity
and $\mymatrix J_{d}$ is the distance Jacobian related to the obstacle
\cite{Marinho2019}. The lower is $\eta_{d}$, the lower is the allowed
approach velocity.

In the case of our example, to prevent collisions between the robot
and the wall, we let $d_{\mathrm{safe}}\triangleq R_{3}$ (see Fig.~\ref{fig:A-more-complete-example})
such that $\tilde{d}\left(t\right)=0\iff d\left(t\right)=R_{3}$,
and the distance function is defined as the distance between $\quat p_{3}$,
which is the center of the circle enclosing the robot, and the wall
plane $\dq{\pi}$. Analogously, to prevent collisions between the
robot and the static cylinders, we use one inequality such as \eqref{eq:Constraint}
for each cylinder, where $d_{\mathrm{safe}}\triangleq R_{i}+R_{3}$
and $d(t)$ is defined as the distance between $\quat p_{3}$ and
the lines $\dq l_{i}$, for $i\in\{1,2\}$. Although those distance
functions and corresponding Jacobians can be computed by using the
classes \texttt{DQ\_Geometry} and \texttt{DQ\_Kinematics}, respectively,
details of how they are obtained using dual quaternion algebra are
presented in \cite{Marinho2019}.

\begin{figure}[tbh]
\noindent \begin{centering}
\subfloat[Circle partially drawn at $t=36\unit{s}$.\label{fig:manipulator-36s}]{\noindent \begin{centering}
\includegraphics[width=1\columnwidth]{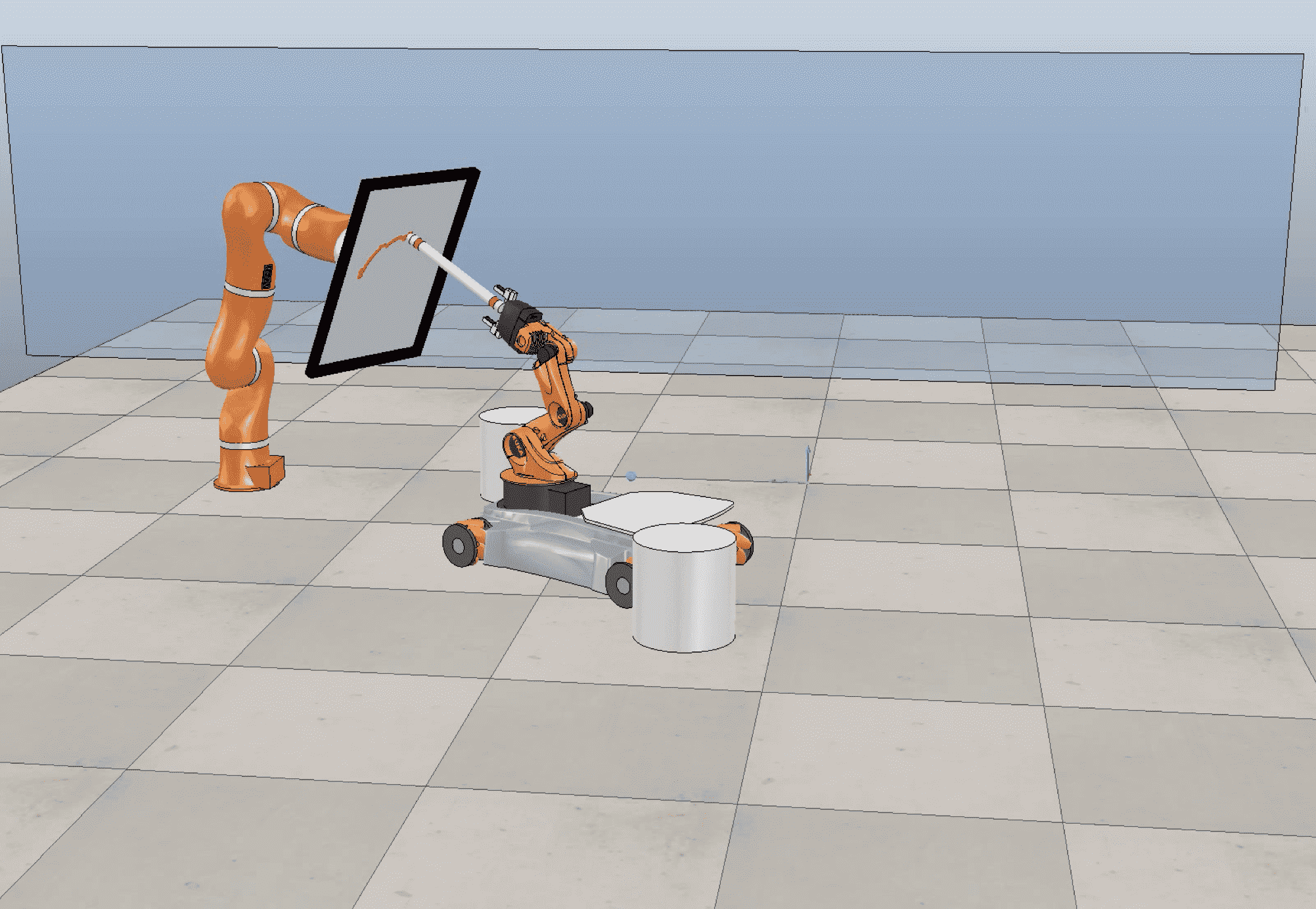}
\par\end{centering}
}
\par\end{centering}
\noindent \begin{centering}
\subfloat[Circle completely drawn at $t=170\unit{s}$.\label{fig:manipulator-170s}]{\noindent \begin{centering}
\includegraphics[width=1\columnwidth]{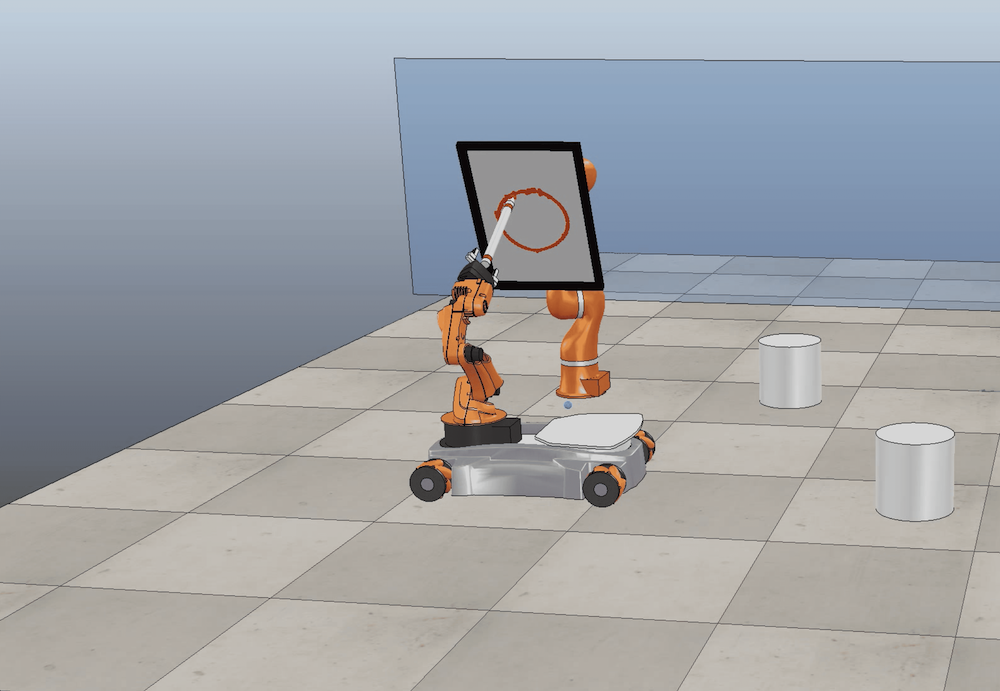}
\par\end{centering}
}
\par\end{centering}
\noindent \begin{centering}
\subfloat[Top view showing the mobile manipulator trajectory. \label{fig:top-view-trajectory}]{\noindent \begin{centering}
\includegraphics[width=1\columnwidth]{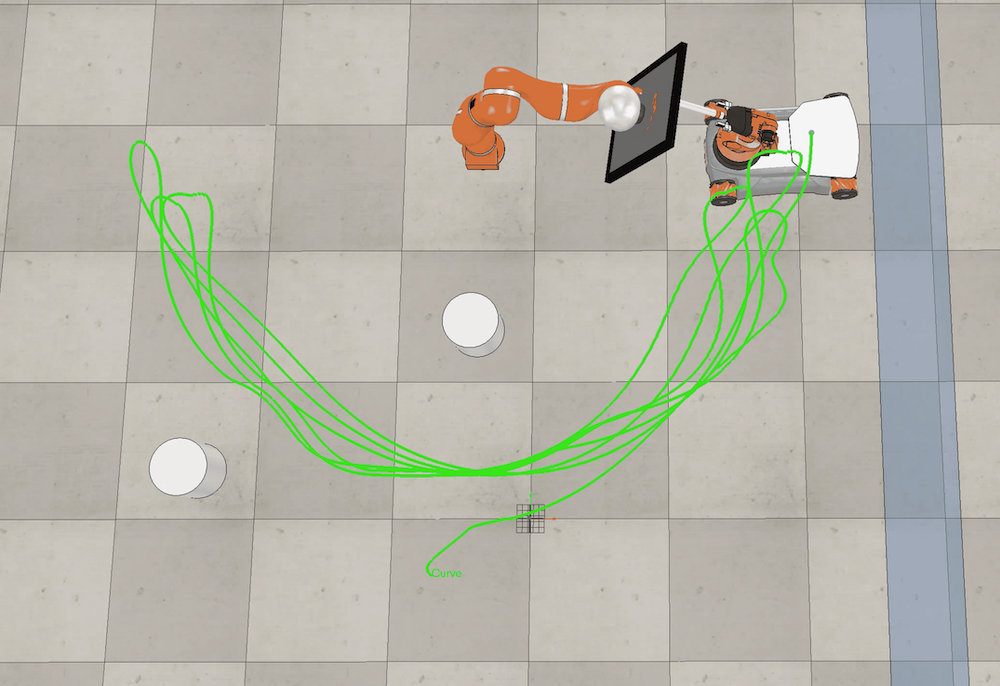}
\par\end{centering}
}
\par\end{centering}
\caption{Manipulator robot interacting with a mobile manipulator in a workspace
with two cylindrical obstacles and a wall. The parameters for the
trajectory generation were $\omega_{n}=0.1$, $\omega_{d}=0.5$ and
$d_{z}=0.1$.\label{fig:VREP-scene}}
\end{figure}

Whenever the felt-tip is sufficiently close to the whiteboard, it
starts to draw a circle. Figs.~\ref{fig:manipulator-36s} and \ref{fig:manipulator-170s}
show that the mobile manipulator successfully draws the circle, although
there are some small imperfections as the obstacle avoidance is a
hard constraint that is always respected at the expense of worse trajectory
tracking. Last, Fig.~\ref{fig:top-view-trajectory} shows the mobile
manipulator trajectory, which is always adapted to prevent collisions
with the cylinders and the wall.

The main MATLAB code is shown in Listing~\ref{lis:Complete-Matlab-code}.
For the sake of conciseness, the functions \texttt{compute\_lwr4\_reference}
and \texttt{compute\_youbot\_reference}, which calculate the aforementioned
end-effector trajectories, are omitted. The same applies for \texttt{get\_plane\_from\_vrep
}and\texttt{ get\_line\_from\_vrep}, which get information about the
location of the corresponding geometric primitives in V-REP, and\texttt{
compute\_constraints}, which calculates constraints such as \eqref{eq:Constraint}
for each obstacle in the scene. The original source code used in this
example is available as supplementary material. For a more up-to-date
version, refer to \url{https://github.com/dqrobotics/matlab-examples}.

\begin{lstlisting}[caption={Main MATLAB code for the simulation.},label={lis:Complete-Matlab-code},float,basicstyle={\scriptsize\ttfamily},tabsize=4]
sampling_time = 0.05;
total_time = 200;
for t=0:sampling_time:total_time

%% Get obstacles from V-REP          
plane = get_plane_from_vrep(vi,'ObstaclePlane',DQ.k);     
cylinder1 = get_line_from_vrep(vi,'ObstacleCylinder1',DQ.k);     
cylinder2 = get_line_from_vrep(vi,'ObstacleCylinder2',DQ.k);         
	
%% Set references for both robots     
[lwr4_xd, lwr4_ff] = compute_lwr4_reference(lwr4,...
	simulation_parameters, lwr4_x0, t);         
[youbot_xd, youbot_ff] = compute_youbot_reference(...
	youbot_control, lwr4_xd, lwr4_ff);
	
%% Compute the control input for the manipulator  
lwr4_u = lwr4_controller.compute_tracking_control_signal(...
	lwr4_q, vec8(lwr4_xd),vec8(lwr4_ff));       

%% Compute constrained control input for the youbot
[Jconstraint, bconstraint] = compute_constraints(youbot, ...
	youbot_q, plane,cylinder1,cylinder2);   
youbot_control.set_inequality_constraint(-Jconstraint,...
	1*bconstraint);
youbot_u= youbot_control.compute_tracking_control_signal(...
	youbot_q, vec8(youbot_xd), vec8(youbot_ff));
	
% Since we are using V-REP just for visualization, integrate 
% the control signal to update the robots configurations     
lwr4_q = lwr4_q + sampling_time*lwr4_u;  
youbot_q = youbot_q + sampling_time*youbot_u;
	
%% Send desired values to V-REP   
lwr4_vreprobot.send_q_to_vrep(lwr4_q);
youbot_vreprobot.send_q_to_vrep(youbot_q);
end
\end{lstlisting}

\section{\label{sec:development_infrastructure}Development infrastructure}

Aiming at the scalability of the DQ Robotics library and taking advantage
of the familiarity that current developers have with Git and Github,
we opted to also use those services.

End-users are agnostic to these infrastructural decisions. For them,
usage, performance, and installation of the library matter the most.
The usage of the library has been simplified by relying on a programming
syntax similar to the mathematical description and being attentive
to proper object-oriented programming practices. Furthermore, good
performance is achieved by careful optimization of the library and
by the C++ implementation.

The ease of installation has been addressed in language/platform specific
ways that we document in details on the DQ Robotics' Read the Docs.\footnote{https://dqroboticsgithubio.readthedocs.io/en/latest/installation.html}

For MATLAB, DQ Robotics is distributed as a Toolbox that can be installed
by the end-user in any of the MATLAB-compatible operating systems.

For C++, we officially focus on Ubuntu long-term service (LTS) distributions,
in a similar way to the most recent distributions of ROS and other
large open-source libraries such as TensorFlow.\footnote{https://www.tensorflow.org/}
For any Ubuntu LTS distribution that has not reached its end-of-life,\footnote{https://ubuntu.com/about/release-cycle}
the user can install the C++ version of the library with the following
three commands

\begin{lstlisting}[language=sh]
sudo add-apt-repository ppa:dqrobotics-dev/release 
sudo apt-get update 
sudo apt-get install libdqrobotics
\end{lstlisting}
made available via our release Personal Package Archive (PPA),\footnote{https://launchpad.net/\textasciitilde dqrobotics-dev/+archive/ubuntu/release}
in which we store the latest stable versions of the library. Interfaces
between DQ Robotics and other libraries also reside in the same PPA.
For example, after adding the PPA, the user can install the V-REP
interface with the following command

\begin{lstlisting}[language=sh]
sudo apt-get install libdqrobotics-interface-vrep
\end{lstlisting}

The PPA guarantees that the user can reliably install the package
in their system as the PPA only stores packages that compiled successfully.
Support for other operating systems will be driven by user interest,
but should not be a big challenge for most systems since we are using
only CMAKE and Eigen3 as project dependencies, both of which are widely
supported.

Lastly, the Python version of DQ Robotics is made available through
a Pybind11-based\footnote{https://github.com/pybind/pybind11} Python
wrapper of the C++ library. This has three main advantages. First,
there is no need to redevelop and maintain versions of the libraries
in different programming languages, which is highly demanding. Second,
unit-testing code written in Python, which is easier to write and
maintain, automatically validates the C++ library as well. Third,
we can have Python's ease-of-use with C++'s performance for each function.
Although there is a computational overhead when using the Python bindings
(in contrast with directly using the C++ code), it is much lower than
a native Python implementation. A simple example of this can be seen
in Table~\ref{tab:multiplication_computational_time}, in which we
compare the average required time for dual quaternion multiplications
in the same machine.
\begin{table}[tph]
\begin{centering}
\caption{\label{tab:multiplication_computational_time}Computational time of
dual-quaternion multiplications.}
\par\end{centering}
\begin{centering}
\begin{tabular*}{1\columnwidth}{@{\extracolsep{\fill}}ccc}
\hline 
 & Mean {[}$\mu$s{]} & Standard Deviation {[}$\mu$s{]}\tabularnewline
\hline 
\hline 
C++ (gcc 5.4.0) & 0.22 & 0.0149\tabularnewline
\hline 
MATLAB R2019a & 8.94 & 0.1223\tabularnewline
\hline 
Python 3.5.2 Bindings & 0.82 & 0.0242\tabularnewline
\hline 
Python 3.5.2 Native & 31.42 & 0.1646\tabularnewline
\hline 
\end{tabular*}
\par\end{centering}
\medskip{}
{\scriptsize{}{*}Mean and standard deviation of the required time
for one dual-quaternion multiplication. This was calculated from a
thousand sets of a thousand executions on the same Intel Core i9 9900K
Ubuntu 16.04 x64 system and excludes the time required to generate
the random dual quaternions.}{\scriptsize\par}
\end{table}

An Ubuntu LTS user can install the Python3 version of DQ Robotics
from the Python Package Index (PyPI)\footnote{https://pypi.org/}
with a single line of code

\begin{lstlisting}[language=sh]
python3 -m pip install --user dqrobotics
\end{lstlisting}
Lastly, the PyPI package is continuously built from source using TravisCI,\footnote{https://travis-ci.com/}
so that the code is properly compiled and tested before being distributed
to our users.

\section{Applications and comparison with other packages and libraries}

\begin{figure}[tbh]
\noindent \begin{centering}
\subfloat[The bimanual surgical robot SmartArm \cite{marinho2019integration}.\label{fig:Surgical-robots}]{\noindent \begin{centering}
\includegraphics[width=0.86\columnwidth]{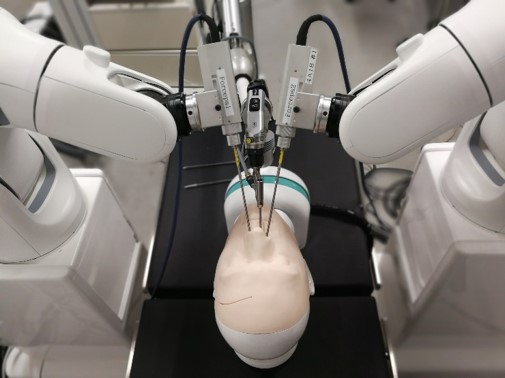}
\par\end{centering}
}
\par\end{centering}
\noindent \begin{centering}
\subfloat[Whole-body control of a humanoid robot \cite{Quiroz-Omana2019}.\label{fig:humanoid}]{\noindent \begin{centering}
\includegraphics[width=0.86\columnwidth]{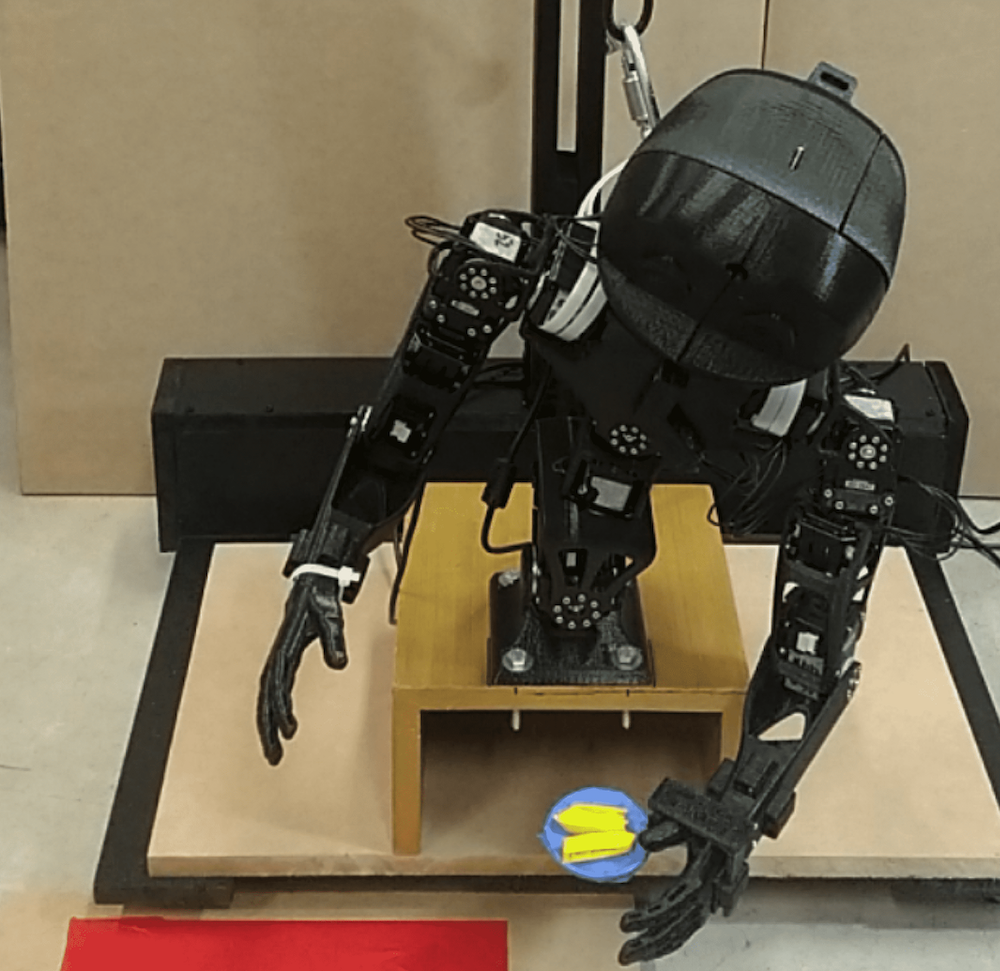}
\par\end{centering}
}
\par\end{centering}
\noindent \begin{centering}
\subfloat[Cooperative mobile manipulators\cite{Savino2018}.\label{fig:cooperative}]{\noindent \begin{centering}
\includegraphics[width=0.86\columnwidth]{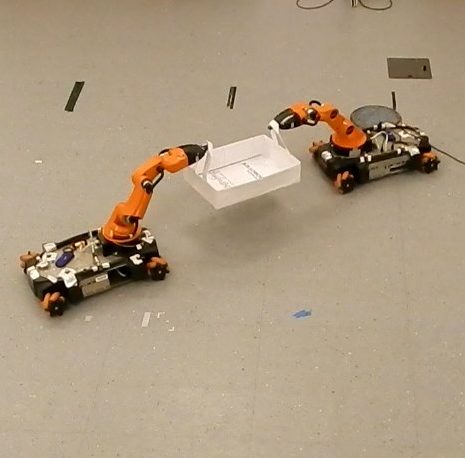}
\par\end{centering}
}
\par\end{centering}
\caption{Examples of different applications of DQ Robotics.\label{fig:applications}}
\end{figure}

DQ Robotics has been used for more than nine years to model and control
different robots in applications such as bimanual surgical robots
in constrained workspaces (Fig.~\ref{fig:Surgical-robots}), whole-body
control of humanoid robots (Fig.~\ref{fig:humanoid}), and the decentralized
formation control of mobile manipulators (Fig.~\ref{fig:cooperative}).\footnote{A more complete list of papers and applications can be found in \url{https://dqrobotics.github.io/citations.html}.}

\noindent 

The main difference from other packages and libraries is that DQ Robotics
implements dual quaternion algebra in a close way to the mathematical
notation. Furthermore, all supported languages, namely Python, MATLAB,
and C++, use the same convention and style and it is straightforward
to port code in one language to other supported languages. To enable
that, we avoid using features that are not available in all three
languages, unless that would lead to poor-quality code. From the user
point-of-view, the advantage of such approach is that users can prototype
their ideas on MATLAB using the convenient plot system, then easily
translate the code to C++ and deploy it on a real robotic platform.
The Python version of the library is somewhere between MATLAB and
C++, being suitable for prototyping and also for deployment on real
applications because C++ runs under the hood.

In terms of target audience, the MATLAB version of DQ Robotics has
some overlap with Peter Corke's Robotics Toolbox \cite{Corke2017},
most notably in terms of their use for education and learning. On
the one hand, earlier versions were greatly inspired by Peter Corke's
Robotics Toolbox, most notably the plot system, which is particularly
useful when teaching robotics. On the other hand, DQ Robotics is entirely
based on general dual quaternion algebra whereas Robotics Toolbox
uses classic representations for robot modeling, although it does
offer support for unit quaternions. Moreover, other functionalities
of Peter Corke's Robotics Toolbox such as path planning, localization,
and mapping are not available in DQ Robotics as of now. However, some
functionalities of Peter Corke's Robotics Toolbox are now available
in toolboxes distributed by MathWorks, such as the Image Processing
Toolbox\footnote{\url{https://www.mathworks.com/products/image.html}}
and the Robotics System Toolbox,\footnote{\url{https://www.mathworks.com/products/robotics.html}}
which are, in principle, compatible with DQ Robotics.

Another MATLAB toolbox \cite{Leclercq2013} focuses on the application
of unit dual quaternions to the neuroscience domain and presents basic
kinematic modeling of serial mechanisms. In addition to having many
fewer functionalities than DQ Robotics, it uses procedural programming,
which makes the code quite different from the notation used on paper.

A major contrast between DQ Robotics and Peter Corke's and Leclercq's
MATLAB toolboxes is that DQ Robotics has Python and C++ versions with
identical APIs, as far as the programming languages permit. This increases
the number of possible users and considerably reduces the time required
to go from prototyping to deployment. Moreover, this makes DQ Robotics
compatible with powerful libraries available in Python such as SciPy\footnote{\url{https://www.scipy.org/}}
and scikit-learn.\footnote{\url{https://scikit-learn.org/stable/}}
In addition, both Python and C++ versions of the library can be readily
used alongside ROS, which is the current standard of shareable code
in the robotics community. Lastly, a modern development infrastructure,
not available in other existing libraries, makes the library scalable
and easier to maintain and extend.

\section{Conclusions}

This paper presented DQ Robotics, a comprehensive computational library
for robot modeling and control using dual quaternion algebra. It has
the unique feature of using a notation very close to the mathematical
description and supporting three different programming languages,
namely MATLAB, Python, and C++, while using very similar conventions
among them, making it very easy to switch between languages and thus
shortening the development time from prototyping to the implementation
on actual robots. The library currently supports serial manipulator
robots, cooperative systems such as two-arm robots, mobile manipulators
and even branched mechanisms such as humanoids. Robot models are generated
automatically from simple geometrical parameters, such as the Denavit-Hartenberg
parameters. In addition, robots can be easily combined to yield more
complex ones while the corresponding forward kinematics and differential
kinematics are also computed automatically and analytically at execution
time. The library also offers a rich set of motion controllers and
it is very simple to implement new ones. Although it does not offer
dynamics modeling yet, it can be easily integrated to existing libraries
for that purpose. However, robot dynamics modeling using dual quaternion
algebra is currently under development and will be integrated into
the library as soon as it becomes mature. Rather than being a competitor
to existing libraries, DQ Robotics aims at complementing them and
helping in the popularization of dual quaternion algebra in the robotics
domain.

\section*{Acknowledgment}

The authors would like to thank all users of the DQ Robotics library
for their valuable feedback and bug reports, specially Juan José Quiroz-Omaña
and other members of the MACRO research group at UFMG.

\balance

\bibliographystyle{IEEEtran}
\bibliography{bib/ram}

\begin{thebibliography}{10}
\providecommand{\url}[1]{#1}
\csname url@samestyle\endcsname
\providecommand{\newblock}{\relax}
\providecommand{\bibinfo}[2]{#2}
\providecommand{\BIBentrySTDinterwordspacing}{\spaceskip=0pt\relax}
\providecommand{\BIBentryALTinterwordstretchfactor}{4}
\providecommand{\BIBentryALTinterwordspacing}{\spaceskip=\fontdimen2\font plus
\BIBentryALTinterwordstretchfactor\fontdimen3\font minus
  \fontdimen4\font\relax}
\providecommand{\BIBforeignlanguage}[2]{{%
\expandafter\ifx\csname l@#1\endcsname\relax
\typeout{** WARNING: IEEEtran.bst: No hyphenation pattern has been}%
\typeout{** loaded for the language `#1'. Using the pattern for}%
\typeout{** the default language instead.}%
\else
\language=\csname l@#1\endcsname
\fi
#2}}
\providecommand{\BIBdecl}{\relax}
\BIBdecl

\bibitem{BottemaRoth:1979}
O.~Bottema and B.~Roth, \emph{{Theoretical kinematics}}.\hskip 1em plus 0.5em
  minus 0.4em\relax North-Holland Publishing Company, 1979, vol.~24.

\bibitem{McCarthy1990}
J.~McCarthy, \emph{{Introduction to theoretical kinematics}}, 1st~ed.\hskip 1em
  plus 0.5em minus 0.4em\relax The MIT Press, 1990.

\bibitem{Adorno2011e}
\BIBentryALTinterwordspacing
B.~V. Adorno, ``{Two-arm Manipulation: From Manipulators to Enhanced
  Human-Robot Collaboration [Contribution {\`{a}} la manipulation {\`{a}} deux
  bras : des manipulateurs {\`{a}} la collaboration homme-robot]},'' PhD
  Dissertation, Universit{\'{e}} Montpellier 2, 2011. [Online]. Available:
  \url{https://tel.archives-ouvertes.fr/tel-00641678/}
\BIBentrySTDinterwordspacing

\bibitem{spong2008robot}
M.~W. Spong and M.~Vidyasagar, \emph{Robot dynamics and control}.\hskip 1em
  plus 0.5em minus 0.4em\relax John Wiley \& Sons, 2008.

\bibitem{siciliano2010robotics}
B.~Siciliano, L.~Sciavicco, L.~Villani, and G.~Oriolo, \emph{Robotics:
  modelling, planning and control}.\hskip 1em plus 0.5em minus 0.4em\relax
  Springer Science \& Business Media, 2010.

\bibitem{siciliano2016springer}
B.~Siciliano and O.~Khatib, \emph{Springer handbook of robotics}.\hskip 1em
  plus 0.5em minus 0.4em\relax Springer, 2016.

\bibitem{Leclercq2013}
\BIBentryALTinterwordspacing
G.~Leclercq, P.~Lef{\`{e}}vre, and G.~Blohm, ``{3D kinematics using dual
  quaternions: theory and applications in neuroscience},'' \emph{Frontiers in
  Behavioral Neuroscience}, vol.~7, no. February, p.~7, jan 2013. [Online].
  Available: \url{http://www.ncbi.nlm.nih.gov/pubmed/23443667
  http://journal.frontiersin.org/article/10.3389/fnbeh.2013.00007/abstract}
\BIBentrySTDinterwordspacing

\bibitem{Adorno2017}
\BIBentryALTinterwordspacing
B.~V. Adorno, ``{Robot Kinematic Modeling and Control Based on Dual Quaternion
  Algebra -- Part I: Fundamentals},'' 2017. [Online]. Available:
  \url{https://hal.archives-ouvertes.fr/hal-01478225v1}
\BIBentrySTDinterwordspacing

\bibitem{quigley2009ros}
M.~Quigley, K.~Conley, B.~Gerkey, J.~Faust, T.~Foote, J.~Leibs, R.~Wheeler, and
  A.~Y. Ng, ``Ros: an open-source robot operating system,'' in \emph{ICRA
  workshop on open source software}, vol.~3, no. 3.2.\hskip 1em plus 0.5em
  minus 0.4em\relax Kobe, Japan, 2009, p.~5.

\bibitem{Rohmer2013}
E.~Rohmer, S.~P.~N. Singh, and M.~Freese, ``V-{REP}: A versatile and scalable
  robot simulation framework,'' in \emph{2013 {IEEE}/{RSJ} International
  Conference on Intelligent Robots and Systems}.\hskip 1em plus 0.5em minus
  0.4em\relax {IEEE}, nov 2013.

\bibitem{Marinho2019}
\BIBentryALTinterwordspacing
M.~M. Marinho, B.~V. Adorno, K.~Harada, and M.~Mitsuishi, ``{Dynamic Active
  Constraints for Surgical Robots Using Vector-Field Inequalities},''
  \emph{IEEE Transactions on Robotics}, vol.~35, no.~5, pp. 1166--1185, oct
  2019. [Online]. Available:
  \url{https://ieeexplore.ieee.org/document/8742769/}
\BIBentrySTDinterwordspacing

\bibitem{marinho2019integration}
M.~M. Marinho, K.~Harada, A.~Morita, and M.~Mitsuishi, ``Smartarm: Integration
  and validation of a versatile surgical robotic system for constrained
  workspaces,'' \emph{The International Journal of Medical Robotics and
  Computer Assisted Surgery}, (in press) 2019.

\bibitem{Quiroz-Omana2019}
\BIBentryALTinterwordspacing
J.~J. Quiroz-Omana and B.~V. Adorno, ``{Whole-Body Control With (Self)
  Collision Avoidance Using Vector Field Inequalities},'' \emph{IEEE Robotics
  and Automation Letters}, vol.~4, no.~4, pp. 4048--4053, oct 2019. [Online].
  Available: \url{https://ieeexplore.ieee.org/document/8763977/}
\BIBentrySTDinterwordspacing

\bibitem{Savino2018}
\BIBentryALTinterwordspacing
H.~J. Savino, L.~C. Pimenta, J.~A. Shah, and B.~V. Adorno, ``{Pose consensus
  based on dual quaternion algebra with application to decentralized formation
  control of mobile manipulators},'' \emph{Journal of the Franklin Institute},
  pp. 1--36, oct 2019. [Online]. Available:
  \url{https://linkinghub.elsevier.com/retrieve/pii/S0016003219307161}
\BIBentrySTDinterwordspacing

\bibitem{Corke2017}
\BIBentryALTinterwordspacing
P.~Corke, \emph{{Robotics, Vision and Control}}, 2nd~ed., ser. Springer Tracts
  in Advanced Robotics.\hskip 1em plus 0.5em minus 0.4em\relax Springer
  International Publishing, 2017, vol. 118. [Online]. Available:
  \url{http://link.springer.com/10.1007/978-3-319-54413-7}
\BIBentrySTDinterwordspacing

\end{thebibliography}

\end{document}